\documentclass[sigconf]{acmart}

\newcommand{\bench}{\textbf{Text2GraphQuery-Bench}}
\AtBeginDocument{
  }

\setcopyright{none}
\copyrightyear{2027}
\acmYear{2027}
\acmDOI{}
\acmConference[KDD '27]{Proceedings of the 33rd ACM SIGKDD Conference on Knowledge Discovery and Data Mining}{August 2027}{TBD}
\acmISBN{}
\renewcommand\footnotetextcopyrightpermission[1]{}

\usepackage{cuted}
\usepackage{caption}
\usepackage{listings}
\usepackage[inline]{enumitem}
\usepackage{balance}
\usepackage{bbm}
\usepackage{cleveref}
\usepackage{amsmath}

\usepackage{booktabs}
\usepackage{multirow}
\usepackage{tabularx}
\usepackage{makecell}
\usepackage{xcolor}
\usepackage{pifont}
\usepackage{makecell}
\usepackage{enumitem}
\usepackage{tikz}
\usepackage[most]{tcolorbox}

\newcommand{\circledchar}[2][gray!70]{
    \tikz[baseline=(char.base)]{
        \node[shape=circle,draw=#1,fill=#1,text=white,inner sep=0.75pt,scale=0.85] (char) {#2};
    }
    }
\begin{document}

\title{Text2GraphQuery-Bench: A Text to Graph Query Benchmark}

\author{
Songlin Lyu$^{1\dagger}$,
Lujie Ban$^{2\dagger}$,
Zihang Wu$^{2\dagger}$,
Tianqi Luo$^{3}$,
Jirong Liu$^{4}$,
Ayoub Moussaid$^{5}$,\\
Oskar van Rest$^{5}$,
Heng Lin$^{6}$,
Chenhao Ma$^{2}$,
Nan Tang$^{3}$,
Shipeng Qi$^{1,7}$,\\
Yongchao Liu$^{1}$,
Zhan Qiu$^{1}$,
Juelu Zhang$^{1}$,
Jiajun Zheng$^{1}$
}
\affiliation{%
  \institution{
  $^{1}$Ant Group \enspace
  $^{2}$The Chinese University of Hong Kong, Shenzhen \enspace
  $^{3}$The Hong Kong University of Science and Technology (Guangzhou) \enspace
  $^{4}$Xi'an Polytechnic University \enspace
  $^{5}$Oracle
  \enspace
  $^{6}$Alibaba Group
  \enspace
  $^{7}$LDBC
  }
  \city{}
  \country{}
}

\renewcommand{\shortauthors}{Lyu et al.}

\begin{abstract}
Graph models are fundamental to data analysis in domains rich with complex relationships. Unlike SQL, which benefits from a relatively unified standard and widespread familiarity, graph query languages are diverse (e.g., Cypher, GQL, SQL/PGQ) and far less familiar to most users, making them significantly harder to learn and use. Text-to-Graph-Query systems address this barrier by translating natural language into executable graph queries, enabling LLMs to serve as natural language interfaces for Graph Database Management Systems (GDBMS). Despite recent progress, existing benchmarks are limited in query language coverage, rely on rigid synthesis pipelines, and lack comprehensive evaluation protocols. We present \bench, the first benchmark covering all mainstream declarative property graph query languages (Cypher, GQL, and SQL/PGQ), with 267,276 \textit{(Question, Graph Query)} pairs across 34 databases and 13 domains. Its construction framework overcomes the rigidity of existing pipelines by supporting both adaptation from heterogeneous resources and domain-aware synthesis with evolutionary query generation, and its Graph-IR-based design enables rapid extension to new graph query languages. The evaluation protocol jointly reports Grammar, GLEU, Similarity, and EX under graph-native difficulty levels, question abstraction levels, and schema aliasing. Experiments on 8 LLMs reveal that: (i)~a significant language gap exists---zero-shot GQL and SQL/PGQ Grammar is far below Cypher, yet few-shot prompting largely recovers it; (ii)~fine-tuning an 8B model on GQL and SQL/PGQ reaches or exceeds zero-shot large models on both Grammar and EX, indicating that unfamiliarity with new graph query languages---rather than model capacity---is the primary barrier; (iii)~as supervision increases, syntax errors recede and the bottleneck shifts---GQL toward aggregation logic, SQL/PGQ toward schema linking and filtering; (iv)~higher question abstraction degrades EX with the bottleneck shifting from syntax to intent-to-schema grounding, while schema aliasing causes only a minimal average EX drop; (v)~EX consistently degrades from Easy to Extra Hard, with Extra Hard remaining a persistent bottleneck even under few-shot.
\end{abstract}

\maketitle

\renewcommand{\thefootnote}{\fnsymbol{footnote}}
\footnotetext[1]{$^{\dagger}$Equal contribution.}

\section{Introduction}
\label{sec:intro}

\begin{table*}[t]
    \centering
    \caption{Comparison of Text2GraphQuery-Bench with existing Text-to-SQL and Text-to-Graph-Query datasets. QAL: question abstraction level. Graph-Native: path/reachability constructs beyond fixed joins.}
    \label{tab:comparison}
    \resizebox{\textwidth}{!}{
    \begin{tabular}{l|cccccccc}
    \toprule
    \textbf{Dataset}
    & \textbf{Source}
    & \textbf{Query Lang.}
    & \textbf{\# QAL}
    & \textbf{\# Example}
    & \textbf{\# (Question, Graph Query) Pair$^{\dagger}$}
    & \textbf{\# DB}
    & \textbf{Graph-Native}
    & \textbf{Difficulty Taxonomy}\\
    \midrule
    Spider~\cite{yu2018spider}
    & Human & SQL & - & 10,181 & 10{,}181  & 200  & \ding{55} & \textcolor{red}{\ding{51}} \\
    
    Spider 2.0~\cite{lei2024spider}
    & Human & SQL & - & 632 & 632 & 213  & \ding{55} & \textcolor{red}{\ding{51}} \\
    
    BIRD~\cite{li2023can}
    & Human & SQL & - & 12,751 & 12{,}751 & 95 & \ding{55} & \textcolor{red}{\ding{51}} \\
    
    \midrule
    Text2Cypher~\cite{ozsoy2025text2cypher}
    & Synthetic & Cypher & - & 44,387 & 44{,}387  & 16 &\textcolor{red}{\ding{51}} & \ding{55} \\
    
    SyntheT2C~\cite{zhong2024synthet2c}
    & Synthetic & Cypher & - & 3,300 & 3{,}300  & 2 & \textcolor{red}{\ding{51}} & \ding{55} \\
    
    GQLBench~\cite{su2026gqlbench}
    & Hybrid & Cypher + GQL & - & \makecell{27,352 / 22,235$^{\S}$} & \makecell{27,352 / 22,235$^{\S}$} & \makecell{311 / 260$^{\S}$} & \textcolor{red}{\ding{51}} & \ding{55} \\
    
    \midrule
    \textbf{Text2GraphQuery-Bench}
    & Hybrid$^{\ast}$
    & Cypher + GQL + SQL/PGQ
    & 3
    & 22,273
    & \textbf{267,276}$^{\ddagger}$
    & 34
    & \textcolor{red}{\ding{51}}
    & \textcolor{red}{\ding{51}} \\
    \bottomrule
    \end{tabular}%
    }
    \begin{flushleft}
    \footnotesize{
     $^{\ast}$LLM-based synthesis and translation from existing benchmarks.
     $^{\dagger}$\#examples $\times$ \#query langs $\times$ \#question annotations per example.\\
     $^{\ddagger}$22{,}273 $\times$ 3 query langs $\times$ 4 question annotations (original + 3 abstraction levels).
     $^{\S}$Dialect-specific counts for Cypher / GQL.
    }
    \end{flushleft}
    \end{table*}

Graph Database Management Systems (GDBMSs) are widely used in domains dominated by complex relationships, such as fraud detection~\cite{henderson2020using,huang2022dgraph,bhardwaj2022investigate}, social network analysis~\cite{cattuto2013time,tjortjis2023graph,nicoara2015hermes}, and supply chain optimization~\cite{rauch2024potential,hong2022graph,almahri2025enhancing}. Unlike relational tables, graph data is queried through paths, reachability, and structural patterns~\cite{antelo2020maximum,zhao2018efficient,abu2015exact,garcia2019ligand,DBLP:journals/pacmmod/GuanZMCHCPD23}, making graph databases difficult for non-expert users to access directly.

\begin{figure}[h]
    \centering
    \includegraphics[width=\linewidth]{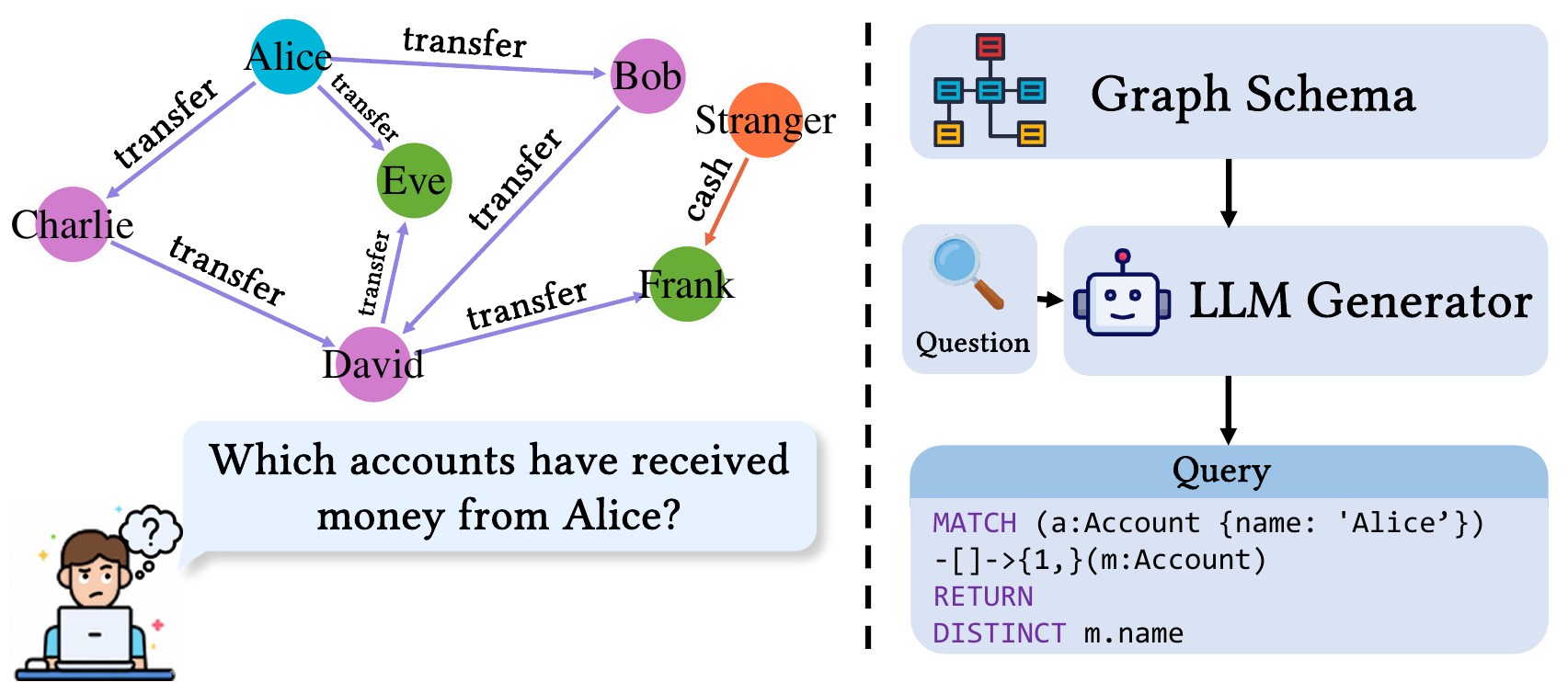}
    \caption{An example of Text-to-Graph-Query task on a financial graph. With the input graph schema and question, the task aims to translate the user intent into a graph query.}
    \label{fig:t2gq_problem}
\end{figure}

Large Language Models (LLMs) have lowered the barrier to structured-data access through natural language interfaces~\cite{liu2024spinach,DBLP:journals/corr/abs-2510-23587,DBLP:journals/tkde/LiuSLMJZFLTL25,mavromatis2025byokg,DBLP:journals/pacmmod/GuanZMCHCPD23,pourrezachase}. Inspired by the success of Text-to-SQL systems~\cite{wang2025agentar,pourrezachase,liu2025xiyan,lialpha,li2025deepeye,DBLP:journals/pvldb/LiLCLT24}, we study \emph{Text-to-Graph-Query}, where a model translates a natural language question and graph schema into an executable graph query, as illustrated in \Cref{fig:t2gq_problem}.

Despite progress in neighboring areas, a standardized benchmark for Text-to-Graph-Query remains missing. Text-to-SQL benchmarks such as Spider~\cite{yu2018spider,lei2024spider} and BIRD~\cite{li2023can,li2025swe} provide reproducible evaluation for relational query generation, while graph-system benchmarks such as LDBC SNB~\cite{szarnyas2022ldbc,erling2015ldbc}, LDBC Financial Benchmark~\cite{qi2023ldbc}, and LDBC Graphalytics~\cite{iosup2016ldbc} expose graph-engine bottlenecks. However, these resources do not evaluate natural-language-to-graph-query generation across languages, schemas, and abstraction levels. Repurposing Text-to-SQL benchmarks is insufficient because SQL is centered on relational operations, whereas graph query languages emphasize path traversal, reachability, and structural constraints; prior SQL-to-graph alignment work such as Graphiti~\cite{he2025graphiti} focuses on query equivalence rather than scalable benchmark construction. Existing Text-to-Graph-Query resources face several coupled limitations:

\noindent\textbf{Limited query language coverage.} Existing benchmarks are confined to a single graph query language—e.g., Text2Cypher~\cite{ozsoy2025text2cypher} and SyntheT2C~\cite{zhong2024synthet2c} support only Cypher, while StockGQL~\cite{liang2024nat} uses only nGQL. In practice, multiple graph query language standards coexist: alongside the widely-used Cypher, GQL—the latest international standard for property graph querying—and SQL/PGQ—introduced in SQL:2023 to enable graph pattern matching within SQL—represent the two newest international standards for graph data access, yet models' ability to generate queries in these emerging standards remains entirely unevaluated.

\noindent\textbf{Rigid data synthesis pipeline.} Existing benchmarks rely on rigid synthesis pipelines that are tightly coupled to specific domains and query templates, making them difficult to generalize to new domains or complex query structures. On the one hand, domain coverage is narrow—e.g., SyntheT2C contains only two databases—and existing pipelines can only generate queries over pre-existing graph databases but cannot synthesize new domain-specific graph database instances grounded in real industrial scenarios. On the other hand, queries are generated via template-based composition—even GQLBench~\cite{su2026gqlbench}, which composes modular templates under schema constraints, cannot produce multi-step \texttt{MATCH} queries or queries with genuine business semantics. Without domain-aware schema synthesis, the pipeline cannot scale to diverse application areas; without an evolutionary query generation mechanism, it cannot produce the structurally complex and semantically meaningful queries needed to stress-test model reasoning on graph-native constructs such as multi-hop path traversal.

\noindent\textbf{Inadequate evaluation protocol.} Text-to-SQL benchmarks define difficulty by the number of JOINs and subquery nesting depth\allowbreak---metrics that fail to capture the topological complexity central to graph queries, where a single \texttt{MATCH} clause with a variable-length path involves no JOINs yet demands multi-hop traversal reasoning absent from SQL complexity measures. Furthermore, prior evaluations rely mainly on execution accuracy, leaving grammar validity and semantic alignment under-characterized—a near-correct query and an incoherent one both score zero, yet represent vastly different capability levels. Real-world users interact with graph databases at vastly different levels of abstraction and vocabulary—an engineer may ask a syntactic-level question explicitly mentioning graph primitives, while a business analyst poses high-level analytical questions with implicit schema assumptions; users also rarely use exact schema element names, instead referring to them via semantically similar aliases. Existing benchmarks consider only a single question abstraction level and original schema names, leaving model robustness to these practical variations entirely unexplored.

These gaps motivate our research question \circledchar[black]{Q}: \textit{How can we design a unified Text-to-Graph-Query benchmark that covers multiple graph query language standards, overcomes the rigidity of existing synthesis pipelines, and enables comprehensive evaluation under diverse difficulty, abstraction, and vocabulary settings?}

\noindent{\textbf{Contributions.}} We present \bench, a unified Text-to-Graph-Query benchmark. \Cref{tab:comparison} positions it against existing datasets. Our contributions are as follows:
\begin{itemize}[leftmargin=*]
    \item We release Text2GraphQuery-Bench, the first benchmark covering all mainstream declarative property graph query languages (Cypher, GQL, and SQL/PGQ), with 267,276 \textit{(Question, Graph Query)} pairs across 34 databases and 13 domains, annotated with question abstraction levels and a graph-native difficulty taxonomy. The dataset, toolkit, and code are available at \url{https://github.com/ldbc/Text2GraphQuery-DataGen}.
    \item We present a flexible construction framework that overcomes the rigidity of existing pipelines: it supports both conversion from heterogeneous resources and domain-aware LLM-based synthesis—generating new graph database instances grounded in industrial scenarios and structurally complex queries via an iterative evolutionary mechanism, with an extensible design for additional graph query languages.
    \item We introduce a comprehensive evaluation protocol that jointly reports \textit{Grammar} validity, Jaro-Winkler-based \textit{Similarity}, \textit{GLEU}-based semantic alignment, and execution accuracy (\textit{EX}) under graph-native difficulty levels, question abstraction levels, and schema aliasing settings.
    \item Experiments on 8 LLMs reveal key insights: (i)~a significant language gap—zero-shot GQL Grammar is 26+ points below Cypher and SQL/PGQ Grammar drops even more sharply (up to 89+ points), yet few-shot prompting largely recovers it; (ii)~fine-tuning an 8B model on GQL and SQL/PGQ reaches or exceeds zero-shot large models on both Grammar and EX (e.g., GQL Grammar 90.8\% vs.\ Claude Opus 4.8 70.0\%, GQL EX 45.1\% vs.\ Gemini-3.5-Flash 48.4\%, SQL/PGQ Grammar 92.1\% vs.\ Claude Opus 4.8 76.1\%, SQL/PGQ EX 40.1\% vs.\ Claude Opus 4.8 2.9\%), indicating that unfamiliarity with new graph query languages—rather than model capacity—is the primary barrier; (iii)~performance degrades as question abstraction increases, with the bottleneck shifting from syntax to intent-to-schema grounding; (iv)~schema aliasing causes only a minimal average EX drop ($-1.26\%$), demonstrating strong model ability to resolve vocabulary variations; (v)~EX consistently degrades from Easy to Extra Hard, with Extra Hard remaining a persistent bottleneck even under few-shot.
\end{itemize}
\section{Related Work}
\label{sec:related_work}

\begin{figure*}[t]
    \centering
    \includegraphics[width=\linewidth]{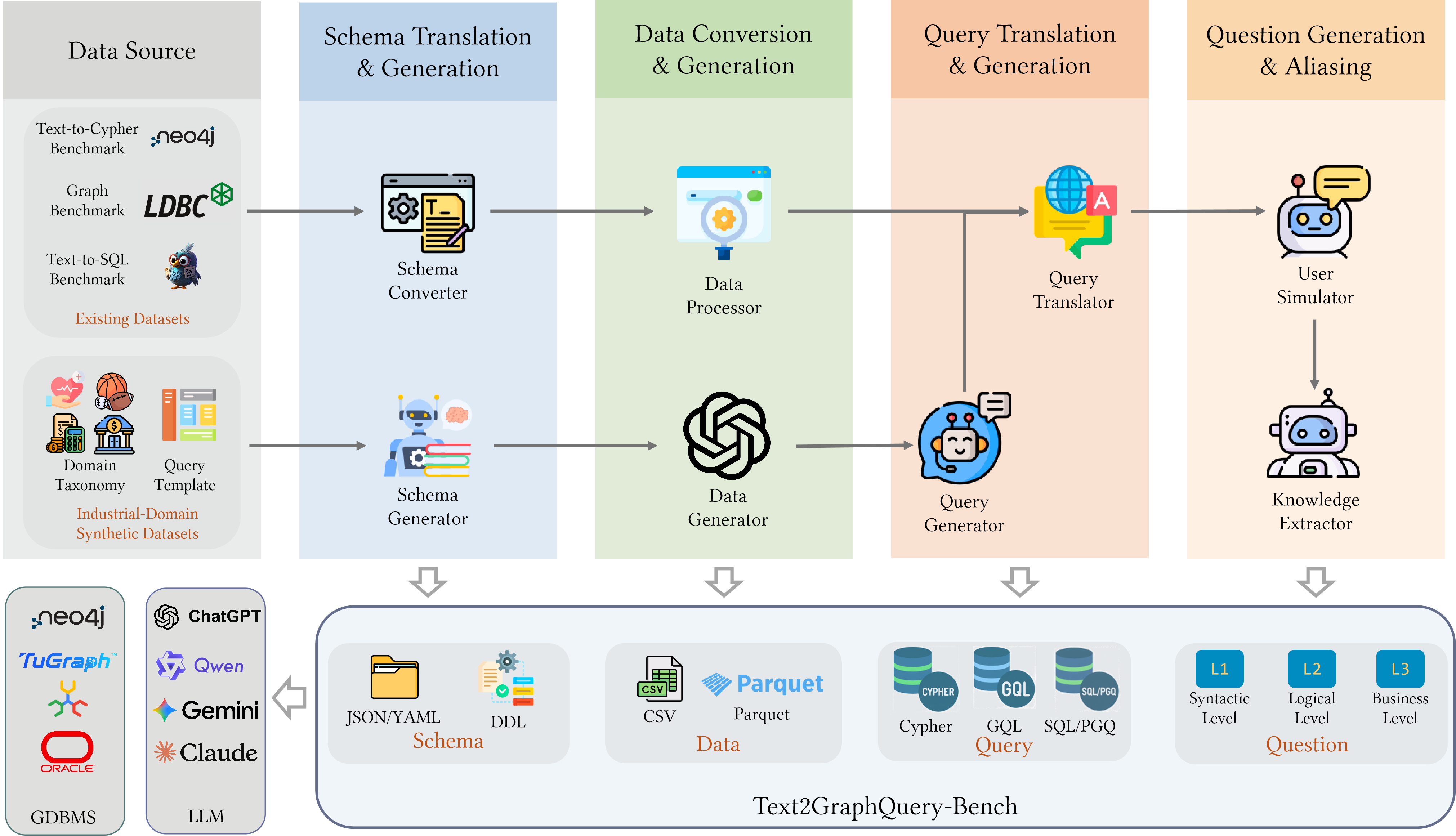}
    \caption{Framework Overview. The 4-stage dataset construction pipeline: (i) Schema Translation \& Generation, (ii) Data Conversion \& Generation, (iii) Query Translation \& Generation, (iv) Question Generation \& Aliasing.}
    \label{fig:framework}
\end{figure*}

\noindent\textbf{Text-to-SQL Benchmark}
Text-to-SQL benchmarks such as Spider~\cite{yu2018spider,lei2024spider} and BIRD~\cite{li2023can,li2025swe} provide reproducible evaluation for relational query generation and have driven continuous improvement in LLM translation abilities~\cite{chen2024tablerag,hong2024knowledge,liu2024mftcoder,li2025deepeye}. However, these benchmarks are centered on join-based relational semantics and SQL operators, and cannot be directly repurposed for Text-to-Graph-Query, which requires graph-native constructs such as path patterns and reachability.

\noindent\textbf{Text-to-Graph-Query Benchmark}
Existing Text-to-Graph-Query benchmarks suffer from three coupled limitations. First, many focus on a single graph query language—Text2Cypher~\cite{ozsoy2025text2cypher}, SyntheT2C~\cite{zhong2024synthet2c}, and CypherBench~\cite{feng2025cypherbench} are confined to Cypher, while nGQL-based datasets~\cite{zhou-etal-2024-r3,liang2024nat,10.1145/3627673.3679713} target only nGQL—precluding cross-dialect evaluation. More recently, GQLBench~\cite{su2026gqlbench} extends cross-domain evaluation to both Cypher and GQL, whereas our benchmark additionally covers SQL/PGQ and evaluates graph-native difficulty, question abstraction, and schema aliasing. Second, their generation processes are tied to specific templates or prompt scaffolds, limiting expansion to diverse schema patterns. Third, their corpora cover only a small number of domains and schemas (e.g., SyntheT2C contains only two databases), limiting cross-domain generalization. These limitations motivate a unified benchmark that scales to diverse domains, supports multiple graph query language standards, and enables standardized, multi-metric evaluation.

\section{Preliminary}
\label{sec:preliminary}

\noindent{\textbf{Graph Database Management System (GDBMS).}} A database system that natively stores and manages data as a graph, prioritizing relationships over relational tables.

\noindent{\textbf{Graph Query Language.}} A domain-specific language for querying graph-structured data. Representative standards include \emph{Cypher}, the widely-adopted declarative language; \emph{ISO-GQL (GQL)}, the new international standard for property graph querying that integrates ideas from Cypher, PGQL, GSQL, and G-CORE; and \emph{ISO-SQL/PGQ (SQL/PGQ)}, introduced in SQL:2023 to enable graph pattern matching within SQL via path patterns and reachability constructs.

\noindent{\textbf{Text-to-Graph-Query.}} Given a natural language question $Q$ and a graph database $D = (\mathcal{G}, \mathcal{S})$ where $\mathcal{G}$ is the graph data and $\mathcal{S}$ is the schema, Text-to-Graph-Query aims to generate an executable graph query $Y = f(Q, D)$ that retrieves the correct answer from $D$.
\section{Benchmark Construction}
\label{sec:bench_construct}

\subsection{Framework Overview}
\label{subsec:datasource}
As shown in \Cref{fig:framework}, our construction framework standardizes raw resources into Text-to-Graph-Query data through four stages: schema translation \& generation, data conversion \& generation, query translation \& generation, and question generation \& aliasing. The pipeline supports both adaptation from existing datasets and synthesis for new domains, ensuring domain scalability and producing executable graph databases with aligned \textit{(Question, Graph Query)} pairs across graph query languages and abstraction levels. The evolutionary query generation stage overcomes template-based limitations to yield queries with genuine business semantics and graph-native structural complexity, while the Graph-IR-based translation design enables rapid extension to new languages.

\subsection{Data Collection}
We collect data from two complementary sources: (i) existing datasets and (ii) newly synthesized industrial domain datasets.

\noindent\textbf{Existing Datasets.}
We incorporate representative datasets from both graph and relational domains. For graph query patterns, we use the Neo4j Text2Cypher dataset~\cite{ozsoy2025text2cypher} (10 databases, 4 domains; Cypher/Neo4j only) and FinBench~\cite{qi2023ldbc}, which provides complex Cypher and GQL queries from financial applications. From the relational side, we select BIRD~\cite{li2023can} (9 databases, 7 domains), whose rich query logic provides valuable reference after relational-to-graph translation.

\noindent\textbf{Industrial-Domain Synthetic Datasets.}
To extend coverage to long-tail domains, we construct synthetic datasets via controlled LLM-based synthesis guided by industrial domain taxonomies, with seed queries abstracted from real business scenarios, generating 14 graph databases in 5 domains.

\subsection{Dataset Construction Framework}
\label{subsec:dsp}
\noindent{\textbf{Schema Translation and Generation.}}
We unify heterogeneous sources via a canonical intermediate representation, \texttt{SchemaGraph}, capturing entities, relations, and property constraints in a model-agnostic manner. For existing relational or graph databases, predefined schemas are parsed into \texttt{SchemaGraph}, where relational artifacts such as associative tables are mapped to edge types with properties. For industrial-domain synthesis datasets, an LLM generates schema specifications under explicit structural constraints, which are then parsed and validated as \texttt{SchemaGraph}.

\noindent{\textbf{Data Conversion and Generation.}}
\label{subsec:data_query_gen}
After schemas are finalized, existing datasets are converted through a deterministic pipeline that normalizes formats and generates import configurations for target graph engines. For industrial-domain schemas without pre-existing instances, we use code-mediated synthesis: the LLM generates executable Python programs conditioned on the \texttt{SchemaGraph} and distribution settings, constructing graph data with controlled scale, referential integrity, and realistic distributions.

\begin{figure}[h]
    \centering
    \includegraphics[width=\linewidth]{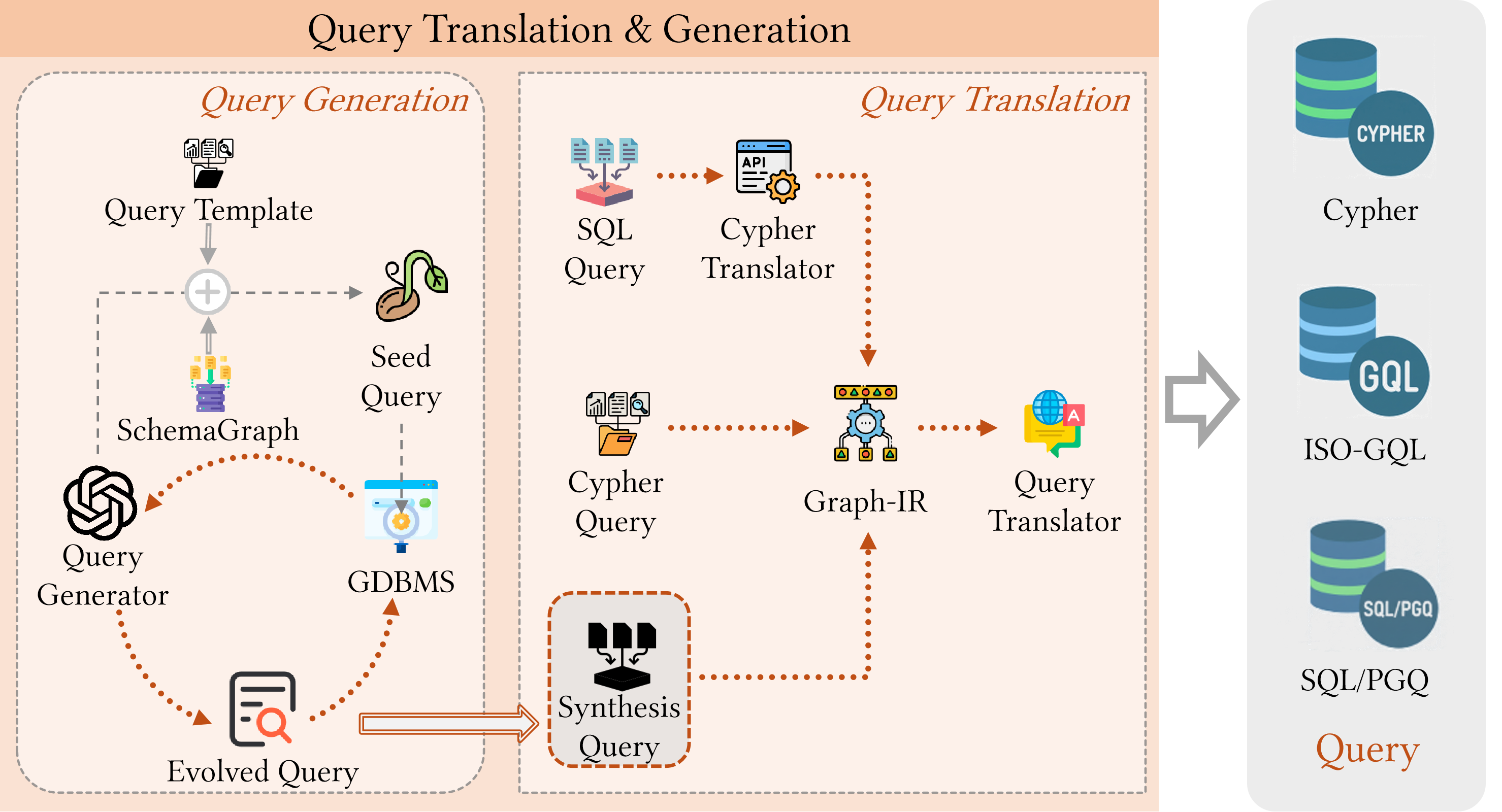}
    \caption{Query Translation and Generation Pipeline.}
    \label{fig:query_trans}
\end{figure}

\noindent{\textbf{Query Translation and Generation.}}
\label{subsec:question_gen}
We construct executable graph queries through a hybrid pathway, as illustrated in \autoref{fig:query_trans}: a multi-stage translator that migrates existing SQL/Cypher logic, and an iterative evolutionary synthesizer that creates new graph queries.

\begin{itemize}[leftmargin=*]
\item \textbf{Query Translation for Existing Datasets.} For existing graph datasets (e.g., FinBench, Text2Cypher) that use Cypher, we design a \texttt{Graph-IR} to represent the basic semantic structure of property graph query languages: we parse Cypher AST into Graph-IR, then translate Graph-IR into GQL and SQL/PGQ. For SQL queries from Text-to-SQL benchmarks (e.g., BIRD), we first convert them into Cypher via open-source tools~\cite{neo4j-s2cypher}—mapping relational operations such as \texttt{JOIN ON} into graph edge traversals—then apply the same Graph-IR pipeline to produce GQL and SQL/PGQ.

\item \textbf{Query Generation for Synthetic Datasets.}
Translation preserves existing query logic but does not ensure sufficient coverage of complex patterns such as 3+ hop traversals, variable-length paths, and multi-step \texttt{MATCH}. Our iterative generation employs a \texttt{dynamic evolution mechanism}: starting from seed graph queries, we execute them on a GDBMS to obtain actual results, then randomly sample queries along with their execution results as grounding context. The LLM is prompted to combine multiple sampled queries—merging their semantics and result patterns—into more complex queries while ensuring the newly generated queries retain genuine business semantics. This execution-validated cycle repeats, progressively transforming simple retrievals into sophisticated reasoning tasks.
\end{itemize}

\noindent{\textbf{Question Generation \& Aliasing.}}
\label{subsec:question_aliasing}
We generate questions along hierarchical abstraction levels with schema aliasing as an additional vocabulary challenge. Each query is annotated at three levels~\cite{lundgard2021accessible}: \textbf{Syntactic Level} questions explicitly mention graph query primitives; \textbf{Logical Level} questions preserve entity-relation logic while abstracting away keywords; and \textbf{Business Level} questions express high-level analytical intent, for which we further derive external knowledge by reverse-rationalizing aligned question-query pairs to make implicit schema or domain assumptions explicit. Additionally, we replace schema element names in questions with semantically similar aliases generated by an LLM; an alias is accepted only if its embedding's closest match in the full schema remains the original entity.

\subsection{Quality Assurance}
We combine automated and human verification to ensure benchmark reliability. All generated queries undergo execution-centric validation in Cypher—chosen for its mature GDBMS support, since queries translated to newer languages (e.g., GQL, SQL/PGQ) may fail due to incomplete GDBMS support rather than query defects—and failed queries are discarded. We also apply semantic overlap filtering, removing one of two \textit{(Question, Graph Query)} pairs when the embedding similarity between either their questions or queries exceeds 0.98. For the test set, we sample 700 examples (18.3\%) for expert evaluation: each instance is scored independently by an expert and a model on two criteria (1--5 scale)—\textit{question--query semantic alignment} and \textit{abstraction level recognition}—with disagreements resolved by a third expert. The overall pass rate reaches 92.9\% (\Cref{tab:query_stat}), confirming pipeline reliability. The training set primarily relies on automated execution validation for scalability.

\subsection{Difficulty Definition}
\label{subsec:difficulty}
We organize the verified corpus into a graph-native four-tier difficulty taxonomy calibrated around topological and logical complexity: \textbf{Easy} queries cover single-node or single-edge patterns without aggregation or complex filtering; \textbf{Medium} queries add one-hop traversal, simple aggregation (e.g., \texttt{COUNT}, \texttt{SUM}), or basic filters without nesting; \textbf{Hard} queries involve multi-hop paths ($\leq$2 hops or variable-length), multiple conditions, or non-nested aggregation; and \textbf{Extra Hard} queries require complex paths ($\geq$3 hops), multi-step \texttt{MATCH} clauses, nested aggregation, or high structural and logical depth. Together with question abstraction levels and schema aliasing, this taxonomy defines the evaluation dimensions of our benchmark.
\begin{table}[h]
    \centering
    \caption{Overview of the Generated Dataset Statistics.}
    \label{tab:dataset_overview}
    \resizebox{\linewidth}{!}{
    \renewcommand{\arraystretch}{1.35}
    \begin{tabular}{lcccc}
        \toprule
        \textbf{Metric} & \textbf{Total} & \textbf{Train Set} & \textbf{Dev Set} & \textbf{Test Set} \\
        \midrule
        \# Domain & 13 & 4 & 10 & 5 \\
        \# Database & 34 & 10 & 18 & 6 \\
        \# Example & 22,273 & 12,861 & 5,585 & 3,827 \\
        \# Node Types & 8\,\textcolor{gray}{\footnotesize[1--22]} & 4\,\textcolor{gray}{\footnotesize[1--13]} & 8\,\textcolor{gray}{\footnotesize[4--22]} & 8\,\textcolor{gray}{\footnotesize[7--9]} \\
        \# Edge Types & 7\,\textcolor{gray}{\footnotesize[1--16]} & 4\,\textcolor{gray}{\footnotesize[1--12]} & 7\,\textcolor{gray}{\footnotesize[1--16]} & 8\,\textcolor{gray}{\footnotesize[7--8]} \\
        \# Properties & 46\,\textcolor{gray}{\footnotesize[9--525]} & 20\,\textcolor{gray}{\footnotesize[9--85]} & 54\,\textcolor{gray}{\footnotesize[25--525]} & 46\,\textcolor{gray}{\footnotesize[43--66]} \\
        \# Records & 17.2K\,\textcolor{gray}{\footnotesize[77--13.6M]} & 40.6K\,\textcolor{gray}{\footnotesize[77--13.6M]} & 17.2K\,\textcolor{gray}{\footnotesize[1.7K--1.1M]} & 7.5K\,\textcolor{gray}{\footnotesize[2.6K--65.0K]} \\
        \bottomrule
    \end{tabular}
    }
\end{table}

\section{Data Statistics}
\label{sec:statistic}

\subsection{Dataset Statistics}
\begin{figure}[h]
    \centering
    \includegraphics[width=\linewidth]{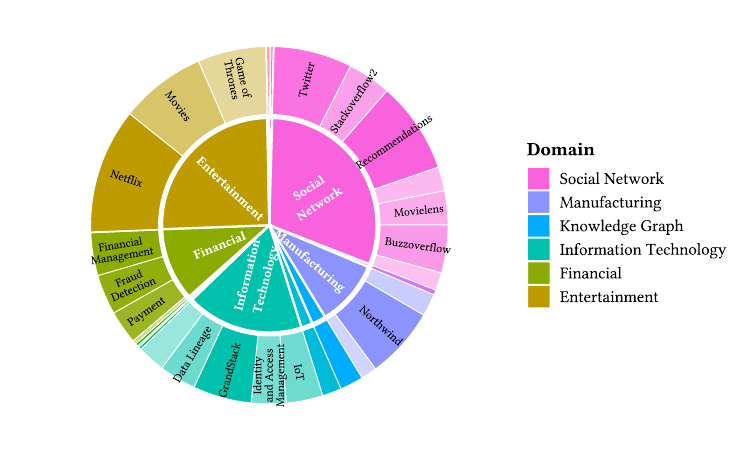}
    \caption{Data Domain Distribution w.r.t. size.}
    \label{fig:dataset_sunburst}
\end{figure}

As shown in \Cref{tab:dataset_overview}, Text2GraphQuery-Bench contains 34 graph databases spanning 13 domains, totaling 22,273 verified examples split into train (10 DBs, 4 domains), development (18 DBs, 10 domains), and test (6 DBs, 5 domains) sets for cross-domain evaluation. Schema and data-size metrics report medians with ranges in brackets. The test set is deliberately constrained to a narrower range (7--9 node types, 2.6K--65.0K records) for evaluation stability, while the dev.\ set retains full diversity. \Cref{fig:dataset_sunburst} shows the domain distribution.

\subsection{Graph Query Statistics}
\label{subsec:gql_stat}
\Cref{tab:query_stat} shows that Medium and Hard queries dominate the benchmark (43.62\% and 30.23\%), and the combined Hard+Extra Hard share reaches 34.99\%, ensuring that over one-third of the benchmark requires multi-hop patterns or nested logic. Average query length increases monotonically with difficulty (86.54$\to$237.24 characters).
Human expert evaluation confirms consistently strong semantic alignment across all difficulty tiers (overall average score 4.81/5, 92.9\% pass rate).
The low rating variance and substantial inter-annotator agreement (overall Fleiss' $\kappa=0.63$, ranging from $0.59$ to $0.68$) further demonstrate that the quality judgments are stable and consistent across difficulty levels.

\begin{table}[h]
    \centering
    \caption{Statistics of Graph Queries Across Difficulty Levels. Var. denotes rating variance between annotators. IAA reports inter-annotator agreement measured by Fleiss' $\kappa$ (agreement beyond chance; higher is better)~\cite{artstein2008agreement}.}
    \label{tab:query_stat}
    \resizebox{\linewidth}{!}{
    \begin{tabular}{lcccccc}
        \toprule
        \textbf{Difficulty} & \textbf{Percentage} & \textbf{Avg. Length} & \textbf{Avg. Score} & \textbf{Pass Ratio} & \textbf{Var.} & \textbf{IAA}  \\
        \midrule
        Easy       & 21.39\% & 86.54 & 4.80 & 96.3\%  & 0.26 & 0.60 \\
        Medium              & 43.62\% & 134.75 & 4.86 & 91.6\%  & 0.50 & 0.64 \\
        Hard                & 30.23\% & 184.25 & 4.80 & 94.3\%  & 0.45 & 0.59 \\
        Extra Hard        & 4.76\% & 237.24 & 4.74 & 89.6\%  & 0.78 & 0.68 \\
        \midrule
        \textbf{Overall} & 100\% & 144.28 & 4.81 & 92.9\%  & 0.51 & 0.63 \\
        \bottomrule
    \end{tabular}}
\end{table}

\subsection{Question Statistics}
We analyze the linguistic properties of questions across abstraction levels.
As summarized in \Cref{tab:question_stat}, \emph{Syntactic Level} questions are the most verbose because they explicitly mention graph query primitives, whereas \emph{Logical Level} and \emph{Business Level} questions are more concise and shift the burden to schema/topology inference; lexical diversity increases markedly at the \emph{Business Level}, reflecting richer domain terminology rather than fixed trigger templates.

Expert evaluation confirms that the intended abstraction levels are reliably recognized (recognition scores $\ge$4.96), with the only notable confusion between Logical Level and Business Level caused by missing specific filter values. Semantic alignment degrades mildly as abstraction strengthens, reflecting an abstraction--faithfulness trade-off.

Each expert spent approximately 2 minutes per instance for quality checks; fully manual construction would take substantially longer, motivating our automated pipeline with post-hoc human verification.

\begin{table}[h]
    \centering
    \caption{Statistic of Questions across Abstraction Levels. Recog. Score denotes whether the question belongs to the target abstraction level. Align. Score denotes whether the question is aligned to the query.}
    \label{tab:question_stat}
    \resizebox{\linewidth}{!}{
    \begin{tabular}{l cccc}
        \toprule
        \multicolumn{1}{c}{\textbf{Abstraction Level}} & \multicolumn{1}{c}{\textbf{Avg. Length}} & \multicolumn{1}{c}{\textbf{Vocab Size}} & \multicolumn{1}{c}{\textbf{Recog. Score}} & \multicolumn{1}{c}{\textbf{Align. Score}} \\
        \midrule
        \textbf{Syntactic Level}  & 40.05 & 1,363 & 5.00 & 4.97 \\
        \textbf{Logical Level}  & 22.60 & 1,422 & 4.97 & 4.89   \\
        \textbf{Business Level}  & 24.24 & 2,577 & 4.96 & 4.81 \\
        \midrule
        \textbf{Original} & 24.17 & 2783 & - & 4.75 \\
        \bottomrule
    \end{tabular}
    }
\end{table}

\begin{table*}[t]
    \centering
    \caption{Multi-Query-Language Performance Comparison across Cypher, GQL, and SQL/PGQ.}
    \label{tab:multi_ql_comparison}
    \resizebox{\textwidth}{!}{
    \begin{tabular}{ll cccc cccc cccc}
        \toprule
        \multirow{2}{*}{\textbf{Model}} & \multirow{2}{*}{\textbf{Strategy}} & \multicolumn{4}{c}{\textbf{Cypher}} & \multicolumn{4}{c}{\textbf{GQL}} & \multicolumn{4}{c}{\textbf{SQL/PGQ}} \\
        \cmidrule(lr){3-6} \cmidrule(lr){7-10} \cmidrule(lr){11-14}
         & & \textbf{Grammar} & \textbf{GLEU} & \textbf{Sim.} & \textbf{EX} & \textbf{Grammar} & \textbf{GLEU} & \textbf{Sim.} & \textbf{EX} & \textbf{Grammar} & \textbf{GLEU} & \textbf{Sim.} & \textbf{EX} \\
        \midrule
        \multicolumn{14}{l}{\textit{\textbf{Large Models}}} \\
        \multirow{2}{*}{Claude Opus 4.8} & Zero-shot & 0.962 & 0.603 & 0.879 & 0.512 & \textsc{0.700} & \textsc{0.620} & \textsc{0.865} & \textsc{0.458} & \textsc{0.761} & \textsc{0.357} & \textsc{0.807} & \textsc{0.029} \\

         & Few-shot & \textbf{0.986} & 0.579 & 0.876 & \textbf{0.576} & \textbf{\textsc{0.840}} & \textsc{0.577} & \textsc{0.861} & \textbf{\textsc{0.582}} & \textsc{0.954} & \textbf{\textsc{0.629}} & \textsc{0.845} & \textsc{0.509} \\
        \midrule
        \multirow{2}{*}{Qwen3.7-Max} & Zero-shot & 0.948 & 0.610 & 0.880 & 0.538 & \textsc{0.672} & \textsc{0.643} & \textsc{0.869} & \textsc{0.469} & \textsc{0.054} & \textsc{0.290} & \textsc{0.775} & \textsc{0.000} \\
         & Few-shot & 0.937 & 0.583 & 0.873 & 0.553 & \textsc{0.798} & \textbf{\textsc{0.650}} & \textbf{\textsc{0.870}} & \textsc{0.548} & \textbf{\textsc{0.962}} & \textsc{0.609} & \textsc{0.821} & \textsc{0.209} \\
        \midrule
        \multirow{2}{*}{GPT-5.5} & Zero-shot & 0.973 & 0.607 & 0.876 & 0.533 & \textsc{0.571} & \textsc{0.629} & \textsc{0.866} & \textsc{0.462} & \textsc{0.403} & \textsc{0.428} & \textsc{0.816} & \textsc{0.015} \\
         & Few-shot & 0.975 & 0.590 & 0.871 & 0.553 & \textsc{0.806} & \textsc{0.603} & \textsc{0.857} & \textsc{0.549} & \textsc{0.928} & \textsc{0.623} & \textbf{\textsc{0.846}} & \textbf{\textsc{0.512}} \\
        \midrule
        \multirow{2}{*}{Gemini-3.5-Flash} & Zero-shot & 0.969 & 0.624 & 0.881 & 0.515 & \textsc{0.662} & \textsc{0.618} & \textsc{0.858} & \textsc{0.484} & \textsc{0.589} & \textsc{0.404} & \textsc{0.813} & \textsc{0.002} \\
         & Few-shot & 0.889 & 0.592 & 0.873 & 0.540 & \textsc{0.735} & \textsc{0.586} & \textsc{0.841} & \textsc{0.536} & \textsc{0.900} & \textsc{0.606} & \textsc{0.821} & \textsc{0.506} \\
        \midrule
        \multirow{2}{*}{DeepSeek-V4-Pro} & Zero-shot & 0.967 & \textbf{0.639} & \textbf{0.891} & 0.480 & \textsc{0.586} & \textsc{0.635} & \textsc{0.865} & \textsc{0.403} & \textsc{0.092} & \textsc{0.292} & \textsc{0.779} & \textsc{0.000} \\
         & Few-shot & 0.979 & 0.629 & 0.885 & 0.491 & \textsc{0.700} & \textsc{0.636} & \textsc{0.863} & \textsc{0.454} & \textsc{0.953} & \textsc{0.560} & \textsc{0.794} & \textsc{0.162} \\
        \midrule
        \multirow{2}{*}{Kimi K2.6} & Zero-shot & 0.968 & 0.602 & 0.881 & 0.529 & 0.639 & 0.522 & 0.861 & 0.453 & \textsc{0.084} & \textsc{0.242} & \textsc{0.757} & \textsc{0.000} \\
         & Few-shot & 0.916 & 0.450 & 0.859 & 0.502 & 0.780 & 0.578 & 0.854 & 0.511 & \textsc{0.941} & \textsc{0.583} & \textsc{0.831} & \textsc{0.272} \\
        \midrule
        \multicolumn{14}{l}{\textit{\textbf{Open Weight Models}}} \\
        \multirow{2}{*}{Qwen3-8B} & Zero-shot & 0.792 & 0.538 & 0.861 &  0.263 & 0.369 & 0.480 & 0.784 & 0.160 & 0.000 & 0.227 & 0.740 & 0.000 \\
         & Fine-tuning & \textbf{0.970} & \textbf{0.584} & \textbf{0.868} &  \textbf{0.358} & \textbf{0.908} & \textbf{0.575} & \textbf{0.848} & \textbf{0.451} & \textbf{0.921} & \textbf{0.628} & \textbf{0.871} & \textbf{0.401} \\
        \midrule
        \multirow{2}{*}{text2cypher-gemma-2-9b} & Zero-shot & 0.638 & 0.528 & 0.860 &  0.184 & 0.266 & 0.515 & 0.816 & 0.134 & 0.214 & 0.152 & 0.671 & 0.000 \\
         & Fine-tuning & - & - & - & - & 0.874 & 0.545 & 0.841 & 0.317 & 0.837 & 0.618 & 0.858 & 0.296 \\
        \bottomrule
    \end{tabular}
    }
\end{table*}

\section{Metrics}
\label{sec:metrics}

We adopt four metrics for comprehensive evaluation.

\noindent{\textbf{Execution Accuracy (EX).}} EX checks whether the predicted query yields the same result as the gold query: $\mathrm{EX} = \frac{1}{N}\sum_{i=1}^{N}\mathbb{I}[ R(\hat{q}_i)=R(q_i^{*}) ]$, where $\mathbb{I}[\cdot]$ is the indicator function; EX is invariant to syntactic variations that preserve semantics.

\noindent{\textbf{Grammar.}} Grammar measures syntactic parsability under the target language: $\mathrm{Grammar} = \frac{1}{N}\sum_{i=1}^{N}\mathbb{I}[\mathrm{Parse}(\hat{q}_i)\neq \varnothing ]$, where $\mathrm{Parse}(\cdot)$ returns a non-empty AST iff parsing succeeds.

\noindent{\textbf{Google-BLEU (GLEU).}} GLEU~\cite{10.3115/1073083.1073135} measures token-level n-gram similarity between $\hat{q}_i$ and $q_i^{*}$ with smoothed precision; we report the corpus-level score.

\noindent{\textbf{Similarity.}} We report Jaro-Winkler similarity~\cite{winkler1990string} $\mathrm{JW}(\hat{q},q)\in[0,1]$ after lightweight normalization to quantify whether the predicted query follows a similar writing style—e.g., naming conventions, clause ordering, and structural patterns—to the gold query.

\section{Experimental Analysis}
\label{sec:exp}
In this section, we conduct a comprehensive evaluation to validate the effectiveness of our proposed framework and answer the following research questions:

\begin{itemize}[leftmargin=*]
    \item \textbf{RQ1 (Multi-Query-Language Performance):} How do models perform across Cypher, GQL, and SQL/PGQ under different strategies, and what does the cross-language gap reveal about the language barrier in Text-to-Graph-Query?
    \item \textbf{RQ2 (Robustness to Question Variations):} How robust are models when questions vary along abstraction levels and schema aliasing?
\end{itemize}

\subsection{Experimental Settings}
\noindent{\textbf{Baseline Models.}} We evaluate our benchmark with a wide range of LLMs, including large frontier models such as Claude Opus 4.8~\cite{anthropic2026claudeopus48}, Qwen3.7-Max~\cite{alibaba2026qwen37max}, GPT-5.5~\cite{openai2026gpt55}, Gemini-3.5-Flash~\cite{google2026gemini35flash}, DeepSeek-V4-Pro~\cite{deepseekai2026deepseekv4}, and Kimi K2.6~\cite{moonshotai2026kimik26}, as well as smaller open-weight baselines like Qwen3-8B~\cite{yang2025qwen3technicalreport} and text2cypher-gemma-2-9b~\cite{ozsoy2025text2cypher}, which has already been fine-tuned on the Text2Cypher dataset for Cypher generation. These models span proprietary and open-weight access, small-scale to large-scale, and general-purpose to code-specific. For models with undisclosed parameter sizes, their performance serves as a point of reference rather than a direct comparison under controlled parameter size conditions.\\
\noindent{\textbf{Strategies.}} We investigate different tuning strategies and In-Context Learning methods:
\begin{enumerate}[leftmargin=*]
    \item \textbf{Zero-shot Prompting:} The model is provided with the target schema (DDL/Graph Schema) and the natural language question, directly generating the graph query without examples. This tests intrinsic generalization.
    \item \textbf{Few-shot In-Context Learning:} We use a fixed $k$-shot prompt ($k=3$). 
Specifically, we prepend the same three demonstrations (Question, Graph Query) pairs to every test instance, selected from the training set and held constant across all evaluations to ensure reproducibility and avoid test-dependent retrieval.
    \item \textbf{Parameter-Efficient Fine-Tuning:} We apply LoRA~\cite{hu2022lora} to fine-tune the open-weights models on our Training Set. 
    We set the LoRA rank $r=16$, alpha $\alpha=32$, and use a learning rate of $2.0 \times 10^{-4}$ with a per-device batch size of 1 and gradient accumulation of 16 steps, for 3 epochs.
 This strategy aims to verify if small models can surpass large models through domain-specific alignment.
\end{enumerate}
\noindent{\textbf{Implementation Details.}} All models are evaluated in non-thinking mode (i.e., without extended chain-of-thought reasoning). All queries are executed and verified on their corresponding database engines: Cypher queries on TuGraph-DB, GQL queries on Spanner Graph, SQL/PGQ queries on Oracle DB, and SQL queries on SQLite.
Experiments for open-weights models were conducted on a server equipped with 1 NVIDIA Tesla V100 GPU (32GB memory).

\noindent{\textbf{Dataset Selection.}} We select all 2,783 examples from the test set that can be executed on TuGraph-DB, Spanner Graph, and Oracle DB to evaluate performance on Cypher, GQL, and SQL/PGQ. The remaining 1,044 examples cannot be executed on Spanner Graph because it does not support the return of graph elements.

\subsection{Multi-Query-Language Performance}
\label{subsec:multi_ql_perf}

To answer \textbf{RQ1}, we evaluate models across Cypher, GQL, and SQL/PGQ under different strategies.
As shown in \Cref{tab:multi_ql_comparison}, a significant language gap exists: even strong large models achieve substantially lower Grammar and EX on GQL and SQL/PGQ than on Cypher in the zero-shot setting (e.g., Qwen3.7-Max: GQL Grammar 0.672 vs.\ Cypher 0.948, SQL/PGQ Grammar 0.054 vs.\ Cypher 0.948), confirming that GQL and SQL/PGQ syntax is unfamiliar to models pre-trained predominantly on Cypher and SQL.
Few-shot prompting largely recovers this gap, bringing GQL Grammar above 0.70 for all large models (above 0.78 for most) and SQL/PGQ Grammar above 0.90 for most models, while GQL EX approaches Cypher levels; however, SQL/PGQ EX recovery is uneven—Claude Opus 4.8 and GPT-5.5 reach 0.509 and 0.512, while Qwen3.7-Max and DeepSeek-V4-Pro remain at 0.209 and 0.162, suggesting that SQL/PGQ's embedding within SQL syntax poses additional challenges beyond in-context grammar transfer.
Notably, fine-tuning enables smaller open-weight models to attain the highest GQL Grammar score (Qwen3-8B: 90.8\%) and approach the GQL EX of much larger proprietary models (45.1\%); on SQL/PGQ, fine-tuned Qwen3-8B even surpasses several few-shot large models (EX 40.1\% vs.\ Qwen3.7-Max 20.9\%, DeepSeek-V4-Pro 16.2\%), indicating that supervision exposing models to the target graph query language and schema grounding is more critical than parameter scale alone.
GLEU and Similarity further reveal the gap between surface resemblance and semantic correctness. GLEU tracks token-level overlap and correlates with Grammar: SQL/PGQ GLEU drops sharply under zero-shot (0.24--0.43) but recovers under few-shot (0.56--0.63), mirroring the Grammar trajectory. However, GLEU is less discriminative than Grammar—Cypher GLEU stays in a narrow band (0.58--0.64) across strategies despite EX varying by up to 6 points—because models still produce many correct tokens even in semantically wrong queries. Similarity is even more stable, remaining above 0.84 for Cypher and GQL across all settings and above 0.74 even for zero-shot SQL/PGQ where Grammar collapses to 0.05. This consistently high Similarity with low EX indicates that models generally adopt a writing style—naming conventions, clause ordering, structural patterns—similar to the gold query even when the semantics diverge, making Similarity a useful indicator of stylistic alignment but a weak proxy for semantic correctness.

\begin{figure}[h]
    \centering
    \includegraphics[width=\linewidth]{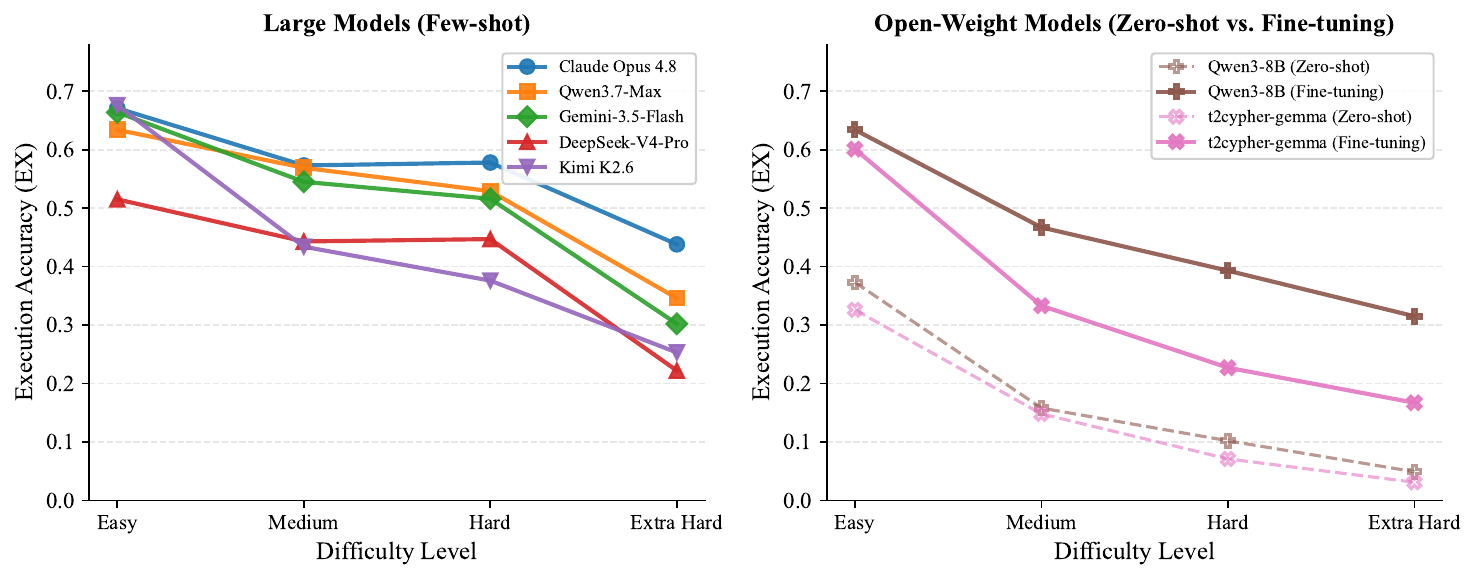}
    \caption{GQL Execution Accuracy (EX) by Difficulty Level. Left: large models under few-shot; Right: open-weight models under zero-shot (dashed) vs.\ fine-tuning (solid).}
    \label{fig:gql_difficulty}
\end{figure}

Performance consistently degrades from Easy to Extra Hard queries, confirming that the benchmark effectively stratifies query difficulty.
As shown in \Cref{fig:gql_difficulty}, we break down GQL EX by difficulty level. Under zero-shot, most models achieve moderate accuracy on Easy queries but collapse on Extra Hard (e.g., GPT-5.5: 0.648$\to$0.191; Qwen3-8B: 0.373$\to$0.049). Few-shot prompting improves all tiers, yet Extra Hard remains a persistent bottleneck—Claude Opus 4.8 reaches only 0.438—suggesting that the complex multi-hop paths, multi-step \texttt{MATCH} clauses, and nested aggregation required by Extra Hard queries demand structural reasoning beyond what in-context examples can instill. Fine-tuning yields the most balanced profile: Qwen3-8B's relative drop from Easy to Extra Hard is nearly halved compared to zero-shot (50.3\% vs.\ 86.9\%), and its Extra Hard EX (0.315) matches few-shot Kimi K2.6 despite being a 100$\times$ smaller model, reinforcing that supervised exposure matters more than parameter scale for structurally complex queries.

\begin{figure}[h]
    \centering
    \includegraphics[width=\linewidth]{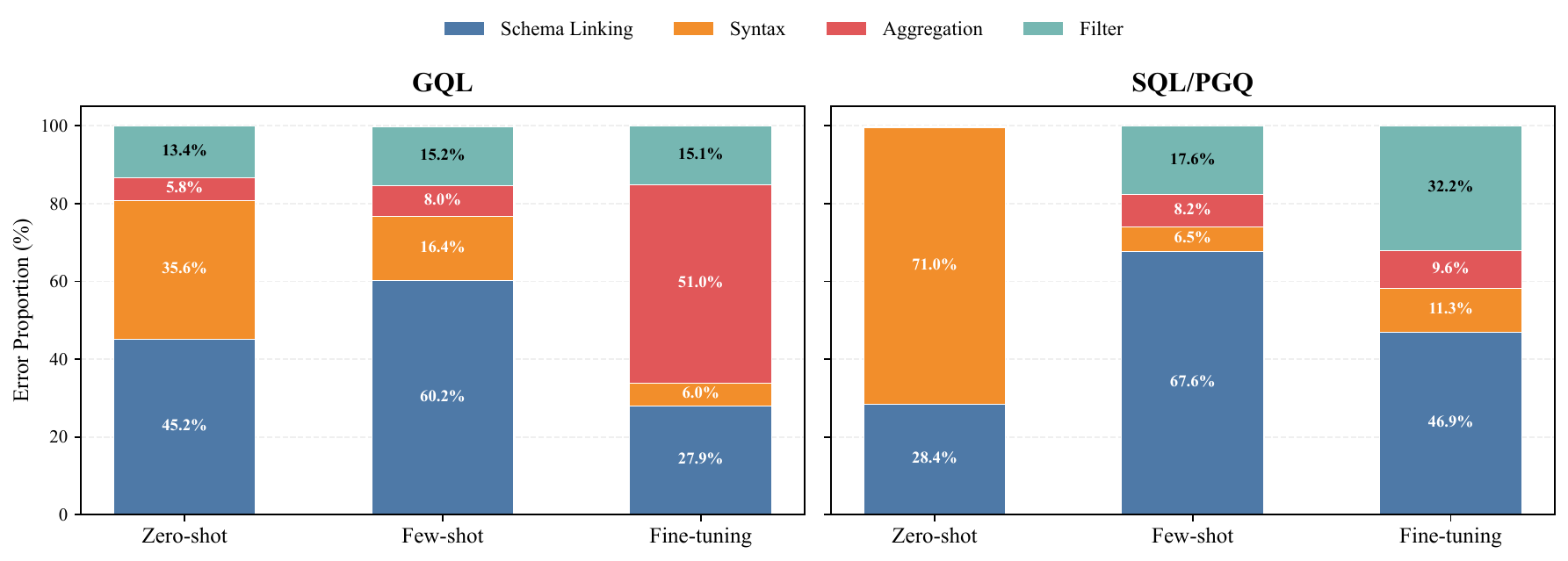}
    \caption{Error distribution comparison between GQL and SQL/PGQ across zero-shot, few-shot, and fine-tuning strategies.}
    \label{fig:failure_analysis}
\end{figure}

We further analyze the error composition across strategies to understand the shifting failure modes. \Cref{fig:failure_analysis} compares the error distribution between GQL and SQL/PGQ under zero-shot, few-shot, and fine-tuning.
A consistent pattern across both languages is that as supervision increases, syntax errors drop sharply from their zero-shot peaks—GQL from 35.6\% to 6.0\%, SQL/PGQ from 71.0\% to 11.3\%—and the error mass shifts toward semantic and logical categories, but the two languages follow distinct trajectories.
In the zero-shot setting, SQL/PGQ failures are overwhelmingly syntax errors (71.0\%), largely because models generate incorrect SQL dialects or omit the \texttt{GRAPH\_TABLE} wrapper, whereas GQL errors are more evenly split between schema linking (45.2\%) and syntax (35.6\%), suggesting that even without GQL exposure, models can partially leverage Cypher-like patterns but struggle with schema grounding.
As syntax errors recede under few-shot prompting, schema linking surges to become the dominant error for both languages (GQL 60.2\%, SQL/PGQ 67.6\%), indicating that few-shot examples effectively convey grammar but provide limited help for mapping natural language to schema elements; notably, GQL syntax errors remain at 16.4\% (vs.\ SQL/PGQ 6.5\%), reflecting GQL's more complex grammar (e.g., \texttt{WHERE} after \texttt{WITH}).
With syntax errors further suppressed by fine-tuning, the two languages diverge sharply: GQL errors concentrate on aggregation (51.0\%), as nested aggregations (e.g., \texttt{AVG(COUNT(t))}) and misuse of \texttt{DISTINCT} become the primary bottleneck, while SQL/PGQ errors shift to schema linking (46.9\%) and filter errors (32.2\%), with aggregation remaining low (9.6\%).
Overall, these trends reveal a shared mechanism—suppression of syntax errors unmasking deeper semantic challenges—yet with language-specific outcomes: GQL's bottleneck shifts from syntax unfamiliarity to aggregation logic, whereas SQL/PGQ's shifts from syntax to schema grounding and condition filtering.

\subsection{Robustness to Question Variations}

To answer \textbf{RQ2}, we evaluate model robustness along the two dimensions introduced in \Cref{subsec:question_gen}: question abstraction levels and schema aliasing.

\begin{figure}[h]
    \centering
    \includegraphics[width=\linewidth]{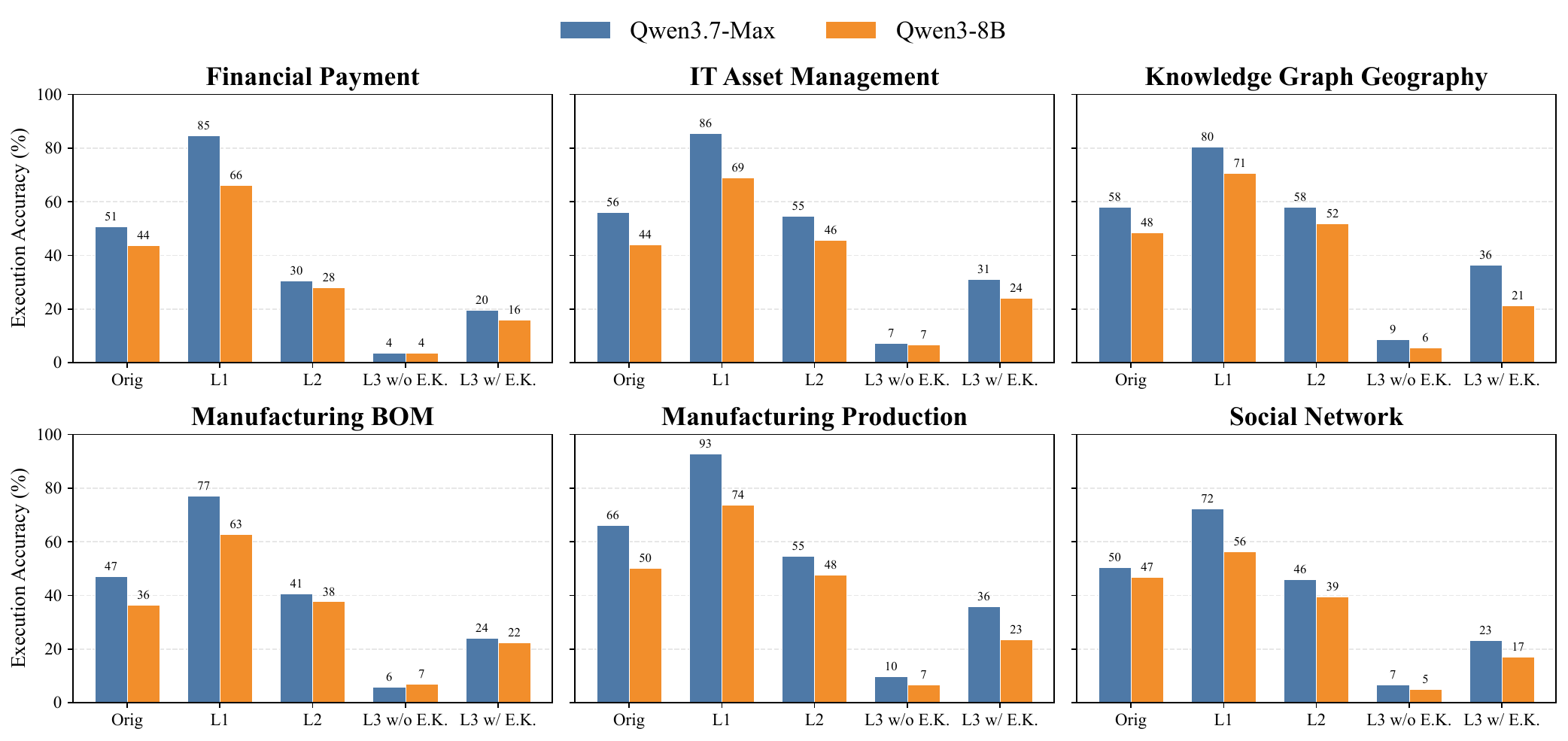}
    \caption{GQL Execution Accuracy by Domain across Question Abstraction Levels. E.K.: External Knowledge.}
    \label{fig:abstraction_analysis}
\end{figure}

\noindent{\textbf{Abstraction Level Sensitivity.}}
We examine robustness to question abstraction with Qwen3.7-Max (few-shot) and Qwen3-8B (fine-tuned).
As shown in \Cref{fig:abstraction_analysis}, Orig.\ denotes the original (unrefined) questions and is closest in abstraction to Logical Level (L2), which typically states the intended logic without explicitly verbalizing graph primitives in Syntactic Level (L1) or fully abstract business intents in Business Level (L3).
Across all domains, both models achieve the highest execution accuracy on L1 (Qwen3.7-Max: 72--93\%; Qwen3-8B: 56--74\%), then degrade as questions move toward L3, with the worst performance occurring at L3 w/o E.K.\ (Qwen3.7-Max: 3.6--9.7\%; Qwen3-8B: 3.6--6.8\%), indicating that the dominant bottleneck at high abstraction is intent-to-schema grounding rather than syntax alone.
Providing external knowledge (E.K.) at L3 w/ E.K.\ consistently recovers accuracy (Qwen3.7-Max: 19.5--36.4\%; Qwen3-8B: 15.9--24.0\%), confirming that explicit grounding signals substantially mitigate analyst-level ambiguity.
Qwen3.7-Max outperforms Qwen3-8B on nearly all domain--level combinations, while their gap narrows sharply at L3 w/o E.K.\ (e.g., Mfg.\ BOM: 5.84\% vs.\ 6.82\%), suggesting that scaling mainly improves syntax/structure alignment whereas ungrounded intent inference remains challenging for both models.

\noindent{\textbf{Schema Aliasing Robustness.}}
We also evaluate whether schema aliasing—replacing schema element names with semantically similar aliases as described in \Cref{subsec:question_gen}—degrades model performance.
As shown in \Cref{tab:alias_robustness}, we compare zero-shot EX on GQL with and without alias using Qwen3.7-Max across six synthesized domains.
The overall trend is that alias increases schema-linking difficulty and reduces accuracy (average $\Delta=-1.26\%$), with the most pronounced drops on Manufacturing Production ($-5.34\%$) and Knowledge Graph Geography ($-2.88\%$).
Interestingly, Financial Payment shows a slight increase, likely because its schema names are semantically intuitive and a few alias rewrites happen to clarify ambiguous expressions; however, this improvement is not statistically significant (paired test $p\approx0.215$), and the remaining five domains all show accuracy declines, confirming that the overall effect of alias is to increase schema-linking difficulty. The average drop is only $-1.26\%$, partly because our alias generation procedure (see \Cref{subsec:question_gen}) constrains each alias to be semantically close to the original schema element; in real-world scenarios, however, both database schema naming itself and user references to schema elements are often more informal and divergent, which would likely cause larger degradation.

\begin{table}[h]
    \centering
    \caption{GQL Zero-shot EX with and without Schema Aliasing (Qwen3.7-Max). $\Delta$ = Alias EX $-$ Original EX.}
    \label{tab:alias_robustness}
    \resizebox{\linewidth}{!}{
    \begin{tabular}{l ccc}
        \toprule
        \textbf{Domain} & \textbf{Original EX} & \textbf{Alias EX} & $\boldsymbol{\Delta}$ \\
        \midrule
        Financial Payment          & 41.18\% & 42.88\% & +1.70\% \\
        IT Asset Management        & 44.84\% & 44.65\% & $-$0.19\% \\
        Knowledge Graph Geography  & 48.78\% & 45.90\% & $-$2.88\% \\
        Manufacturing BOM          & 39.94\% & 39.29\% & $-$0.65\% \\
        Manufacturing Production   & 60.56\% & 55.22\% & $-$5.34\% \\
        Social Network Twitter     & 47.34\% & 45.65\% & $-$1.69\% \\
        \midrule
        \textbf{Average}           & \textbf{46.89\%} & \textbf{45.63\%} & $\boldsymbol{-1.26\%}$ \\
        \bottomrule
    \end{tabular}
    }
\end{table}

\section{Conclusion}
\label{sec:conclusion}
We present Text2GraphQuery-Bench, the first benchmark covering all mainstream declarative property graph query languages (Cypher, GQL, and SQL/PGQ), with 267,276 \textit{(Question, Graph Query)} pairs across 34 databases and 13 domains. Its construction framework overcomes the rigidity of existing pipelines by supporting both adaptation from heterogeneous resources and domain-aware synthesis with evolutionary query generation, with a Graph-IR-based design that enables rapid extension to new languages. To move beyond single-metric evaluation, our protocol assesses both surface quality—Grammar validity, GLEU-based token alignment, and Jaro-Winkler Similarity—and semantic correctness via execution accuracy (EX), reported across graph-native difficulty levels, question abstraction levels, and schema aliasing settings. Experiments on 8 LLMs reveal that the primary bottleneck shifts with supervision: from syntax unfamiliarity under zero-shot, to schema linking under few-shot, to language-specific semantic challenges (aggregation logic for GQL, condition filtering for SQL/PGQ) under fine-tuning. Moreover, while fine-tuning substantially narrows the gap on Easy--Hard queries, Extra Hard accuracy remains low even for the strongest models, identifying nested aggregation and multi-step structural composition as frontier challenges. These findings position Text2GraphQuery-Bench as both a benchmark and an extensible foundation for systematic Text-to-Graph-Query research.
\newpage
\bibliographystyle{ACM-Reference-Format}
\balance
\bibliography{main}
\newpage
\appendix
\onecolumn
\section{Dataset Description}

\subsection{Input Data Format}

Each data entry includes information such as the database name, the original question, the layered reasoning questions, and optional external knowledge.

\captionof{table}{Input data format example.}
\label{tab:input_data_format_example}
\begin{tcolorbox}[
    colback=blue!5!white,
    colframe=blue!75!black,
    title=Input Data Format,
    fonttitle=\bfseries,
    arc=2pt,
    width=\textwidth,
    boxrule=0.5pt,
    left=5pt, right=5pt, top=5pt, bottom=5pt
]
\small
\begin{flushleft}
\texttt{\{} \\
\hspace*{1em} \texttt{"id": "Unique Identifier",} \\
\hspace*{1em} \texttt{"database": "Database Name",} \\
\hspace*{1em} \texttt{"initial\_question": "Original Natural Language Question",} \\
\hspace*{1em} \texttt{"initial\_graph\_query": "Correct Graph Query...",} \\
\hspace*{1em} \texttt{"level\_1": "Level 1 Question",} \\
\hspace*{1em} \texttt{"level\_2": "Level 2 Question",} \\
\hspace*{1em} \texttt{"level\_3": "Level 3 Question",} \\
\hspace*{1em} \texttt{"external\_knowledge": "...",} \\
\hspace*{1em} \texttt{"difficulty": "easy / medium / hard / extra hard",} \\
\hspace*{1em} \texttt{"source": "Data Source"} \\
\texttt{\}}
\end{flushleft}
\end{tcolorbox}

\subsection{Predicted Output Format}

The model generates the corresponding predicted query statement for each data entry.

\captionof{table}{Predicted output format example.}
\label{tab:predicted_output_format_example}
\begin{tcolorbox}[
    colback=blue!5!white,
    colframe=blue!75!black,
    title=Predicted Output Format,
    fonttitle=\bfseries,
    arc=2pt,
    width=\textwidth,
    boxrule=0.5pt,
    left=5pt, right=5pt, top=5pt, bottom=5pt
]
\small
\begin{flushleft}
\texttt{\{} \\
\hspace*{1em} \texttt{"id": "Unique Identifier",} \\
\hspace*{1em} \texttt{"database": "Database Name",} \\
\hspace*{1em} \texttt{"initial\_question": "Original Natural Language Question",} \\
\hspace*{1em} \texttt{"initial\_graph\_query": "Correct Graph Query...",} \\
\hspace*{1em} \texttt{"level\_1": "Level 1 Question",} \\
\hspace*{1em} \texttt{"level\_2": "Level 2 Question",} \\
\hspace*{1em} \texttt{"level\_3": "Level 3 Question",} \\
\hspace*{1em} \texttt{"external\_knowledge": "...",} \\
\hspace*{1em} \texttt{"difficulty": "easy / medium / hard / extra hard",} \\
\hspace*{1em} \texttt{"source": "Data Source"} \\
\hspace*{1em}
\texttt{"initial\_query": "Model Predicted Query Statement",}\\
\hspace*{1em}
\texttt{"level\_1\_query": "Model Predicted Query Statement",}\\
\hspace*{1em}
\texttt{"level\_2\_query": "Model Predicted Query Statement",}\\
\hspace*{1em}
\texttt{"level\_3\_query": "Model Predicted Query Statement",}\\
\hspace*{1em}
\texttt{"level\_3\_external\_knowledge\_query": "Model Predicted Query Statement"}\\
\texttt{\}}
\end{flushleft}
\end{tcolorbox}

\section{Prompts}

\definecolor{appendixprompttitle}{RGB}{239,237,245}
\tcbset{
    appendixprompt/.style={
        enhanced jigsaw,
        breakable,
        colback=white,
        colframe=black,
        colbacktitle=appendixprompttitle,
        coltitle=black,
        fonttitle=\bfseries,
        boxrule=0.6pt,
        titlerule=0.4pt,
        arc=0pt,
        outer arc=0pt,
        width=\textwidth,
        left=6pt,
        right=6pt,
        top=5pt,
        bottom=5pt,
        before skip=6pt,
        after skip=10pt
    }
}
\newcommand{\promptsectiontitle}[1]{
    \par\smallskip
    \noindent\colorbox{appendixprompttitle}{
        \parbox{\dimexpr\linewidth-2\fboxsep\relax}{\strut\bfseries #1}}
    \par\smallskip
}
\newcommand{\promptcaption}[2]{
    \refstepcounter{table}\label{#2}
    \addcontentsline{lot}{table}{\protect\numberline{\thetable}{#1}}
    \begin{center}
        \small\bfseries Table \thetable: #1
    \end{center}
}
\lstset{
    columns=fullflexible,
    breaklines=true,
    breakindent=0pt,
    frame=none,
    keepspaces=true,
    language={},
    showstringspaces=false,
    aboveskip=0pt,
    belowskip=0pt
}

\subsection{Schema Example}

Below we show the schema representation of the Financial Payment domain in three query languages. Although they share the same logical structure (8 vertex types, 8 edge types), the data types and relationship syntax differ across languages because each is instantiated from the same \texttt{SchemaGraph} for a specific target database: openCypher for TuGraph-DB uses TuGraph-native types (\texttt{STRING}, \texttt{DOUBLE}, \texttt{DATETIME}) and Cypher-style relationship patterns; ISO GQL for Spanner Graph uses GQL-standard types (\texttt{STRING(MAX)}, \texttt{FLOAT64}) and GQL-style relationship patterns; and ISO SQL/PGQ for Oracle Database uses SQL types (\texttt{VARCHAR2(4000)}, \texttt{TIMESTAMP}, \texttt{NUMBER(10)}) and SQL/PGQ-style relationship patterns.

\clearpage
\promptcaption{Schema for openCypher.}{tab:cypher_schema}
\begin{tcolorbox}[appendixprompt,title=Schema for openCypher]
\small
\textbf{Vertex types:}
\begin{itemize}
    \setlength\itemsep{0em}
    \item \textbf{Account}: \texttt{account\_id: STRING (REQ)}, \texttt{account\_type: STRING}, \texttt{balance: DOUBLE}, \texttt{currency: STRING}, \texttt{status: STRING}, \texttt{opened\_date: DATE}, \texttt{last\_activity: DATETIME}
    \item \textbf{Customer}: \texttt{customer\_id: STRING (REQ)}, \texttt{customer\_type: STRING}, \texttt{name: STRING}, \texttt{email: STRING}, \texttt{phone: STRING}, \texttt{address: STRING}, \texttt{kyc\_status: STRING}
    \item \textbf{PaymentTransaction}: \texttt{transaction\_id: STRING (REQ)}, \texttt{amount: DOUBLE}, \texttt{currency: STRING}, \texttt{status: STRING}, \texttt{created\_at: DATETIME}, \texttt{completed\_at: DATETIME}, \texttt{reference\_number: STRING}
    \item \textbf{PaymentMethod}: \texttt{method\_id: STRING (REQ)}, \texttt{method\_type: STRING}, \texttt{provider: STRING}, \texttt{status: STRING}, \texttt{expiry\_date: DATE}, \texttt{masked\_identifier: STRING}
    \item \textbf{Merchant}: \texttt{merchant\_id: STRING (REQ)}, \texttt{business\_name: STRING}, \texttt{category\_code: STRING}, \texttt{status: STRING}, \texttt{contract\_terms: STRING}
    \item \textbf{ComplianceRule}: \texttt{rule\_id: STRING (REQ)}, \texttt{jurisdiction: STRING}, \texttt{rule\_type: STRING}, \texttt{effective\_date: DATE}, \texttt{description: STRING}, \texttt{monitoring\_required: BOOL}
    \item \textbf{AuditLog}: \texttt{log\_id: STRING (REQ)}, \texttt{event\_type: STRING}, \texttt{timestamp: DATETIME}, \texttt{details: STRING}
    \item \textbf{RiskAssessment}: \texttt{assessment\_id: STRING (REQ)}, \texttt{score: INT32}, \texttt{assessed\_at: DATETIME}, \texttt{risk\_level: STRING}
\end{itemize}
\smallskip
\textbf{Edge types (all no properties):} Initiates, FundsFrom, ReceivesTo, AuthorizedBy, ProcessedFor, GovernedBy, HasAuditLog, HasRiskAssessment

\smallskip
\textbf{Relationships:} \\
\texttt{(:Customer)-[:Initiates]->(:PaymentTransaction)} \\
\texttt{(:PaymentTransaction)-[:FundsFrom]->(:Account)} \\
\texttt{(:PaymentTransaction)-[:ReceivesTo]->(:Account)} \\
\texttt{(:PaymentTransaction)-[:AuthorizedBy]->(:PaymentMethod)} \\
\texttt{(:PaymentTransaction)-[:ProcessedFor]->(:Merchant)} \\
\texttt{(:PaymentTransaction)-[:GovernedBy]->(:ComplianceRule)} \\
\texttt{(:PaymentTransaction)-[:HasAuditLog]->(:AuditLog)} \\
\texttt{(:PaymentTransaction)-[:HasRiskAssessment]->(:RiskAssessment)}
\end{tcolorbox}

\promptcaption{Schema for ISO GQL.}{tab:gql_schema}
\begin{tcolorbox}[appendixprompt,title=Schema for ISO GQL]
\small
\textbf{Vertex types:}
\begin{itemize}
    \setlength\itemsep{0em}
    \item \textbf{Account}: \texttt{account\_id: STRING(MAX) (REQ)}, \texttt{account\_type: STRING(MAX)}, \texttt{balance: FLOAT64}, \texttt{currency: STRING(MAX)}, \texttt{status: STRING(MAX)}, \texttt{opened\_date: DATE}, \texttt{last\_activity: STRING(MAX)}
    \item \textbf{Customer}: \texttt{customer\_id: STRING(MAX) (REQ)}, \texttt{customer\_type: STRING(MAX)}, \texttt{name: STRING(MAX)}, \texttt{email: STRING(MAX)}, \texttt{phone: STRING(MAX)}, \texttt{address: STRING(MAX)}, \texttt{kyc\_status: STRING(MAX)}
    \item \textbf{PaymentTransaction}: \texttt{transaction\_id: STRING(MAX) (REQ)}, \texttt{amount: FLOAT64}, \texttt{currency: STRING(MAX)}, \texttt{status: STRING(MAX)}, \texttt{created\_at: STRING(MAX)}, \texttt{completed\_at: STRING(MAX)}, \texttt{reference\_number: STRING(MAX)}
    \item \textbf{PaymentMethod}: \texttt{method\_id: STRING(MAX) (REQ)}, \texttt{method\_type: STRING(MAX)}, \texttt{provider: STRING(MAX)}, \texttt{status: STRING(MAX)}, \texttt{expiry\_date: DATE}, \texttt{masked\_identifier: STRING(MAX)}
    \item \textbf{Merchant}: \texttt{merchant\_id: STRING(MAX) (REQ)}, \texttt{business\_name: STRING(MAX)}, \texttt{category\_code: STRING(MAX)}, \texttt{status: STRING(MAX)}, \texttt{contract\_terms: STRING(MAX)}
    \item \textbf{ComplianceRule}: \texttt{rule\_id: STRING(MAX) (REQ)}, \texttt{jurisdiction: STRING(MAX)}, \texttt{rule\_type: STRING(MAX)}, \texttt{effective\_date: DATE}, \texttt{description: STRING(MAX)}, \texttt{monitoring\_required: BOOL}
    \item \textbf{AuditLog}: \texttt{log\_id: STRING(MAX) (REQ)}, \texttt{event\_type: STRING(MAX)}, \texttt{timestamp: STRING(MAX)}, \texttt{details: STRING(MAX)}
    \item \textbf{RiskAssessment}: \texttt{assessment\_id: STRING(MAX) (REQ)}, \texttt{score: INT64}, \texttt{assessed\_at: STRING(MAX)}, \texttt{risk\_level: STRING(MAX)}
\end{itemize}
\smallskip
\textbf{Edge types (all no properties):} Initiates, FundsFrom, ReceivesTo, AuthorizedBy, ProcessedFor, GovernedBy, HasAuditLog, HasRiskAssessment

\smallskip
\textbf{Relationships:} \\
\texttt{(:Customer)-[:Initiates]->(:PaymentTransaction)} \\
\texttt{(:PaymentTransaction)-[:FundsFrom]->(:Account)} \\
\texttt{(:PaymentTransaction)-[:ReceivesTo]->(:Account)} \\
\texttt{(:PaymentTransaction)-[:AuthorizedBy]->(:PaymentMethod)} \\
\texttt{(:PaymentTransaction)-[:ProcessedFor]->(:Merchant)} \\
\texttt{(:PaymentTransaction)-[:GovernedBy]->(:ComplianceRule)} \\
\texttt{(:PaymentTransaction)-[:HasAuditLog]->(:AuditLog)} \\
\texttt{(:PaymentTransaction)-[:HasRiskAssessment]->(:RiskAssessment)}
\end{tcolorbox}

\clearpage
\promptcaption{Schema for ISO SQL/PGQ.}{tab:sql_pgq_schema}
\begin{tcolorbox}[appendixprompt,title=Schema for ISO SQL/PGQ]
\small
\textbf{Vertex types:}
\begin{itemize}
    \setlength\itemsep{0em}
    \item \textbf{"Account"}: \texttt{"account\_id": VARCHAR2(4000) (REQ)}, \texttt{"account\_type": VARCHAR2(4000)}, \texttt{"balance": FLOAT}, \texttt{"currency": VARCHAR2(4000)}, \texttt{"status": VARCHAR2(4000)}, \texttt{"opened\_date": DATE}, \texttt{"last\_activity": TIMESTAMP}
    \item \textbf{"Customer"}: \texttt{"customer\_id": VARCHAR2(4000) (REQ)}, \texttt{"customer\_type": VARCHAR2(4000)}, \texttt{"name": VARCHAR2(4000)}, \texttt{"email": VARCHAR2(4000)}, \texttt{"phone": VARCHAR2(4000)}, \texttt{"address": VARCHAR2(4000)}, \texttt{"kyc\_status": VARCHAR2(4000)}
    \item \textbf{"PaymentTransaction"}: \texttt{"transaction\_id": VARCHAR2(4000) (REQ)}, \texttt{"amount": FLOAT}, \texttt{"currency": VARCHAR2(4000)}, \texttt{"status": VARCHAR2(4000)}, \texttt{"created\_at": TIMESTAMP}, \texttt{"completed\_at": TIMESTAMP}, \texttt{"reference\_number": VARCHAR2(4000)}
    \item \textbf{"PaymentMethod"}: \texttt{"method\_id": VARCHAR2(4000) (REQ)}, \texttt{"method\_type": VARCHAR2(4000)}, \texttt{"provider": VARCHAR2(4000)}, \texttt{"status": VARCHAR2(4000)}, \texttt{"expiry\_date": DATE}, \texttt{"masked\_identifier": VARCHAR2(4000)}
    \item \textbf{"Merchant"}: \texttt{"merchant\_id": VARCHAR2(4000) (REQ)}, \texttt{"business\_name": VARCHAR2(4000)}, \texttt{"category\_code": VARCHAR2(4000)}, \texttt{"status": VARCHAR2(4000)}, \texttt{"contract\_terms": VARCHAR2(4000)}
    \item \textbf{"ComplianceRule"}: \texttt{"rule\_id": VARCHAR2(4000) (REQ)}, \texttt{"jurisdiction": VARCHAR2(4000)}, \texttt{"rule\_type": VARCHAR2(4000)}, \texttt{"effective\_date": DATE}, \texttt{"description": VARCHAR2(4000)}, \texttt{"monitoring\_required": BOOLEAN}
    \item \textbf{"AuditLog"}: \texttt{"log\_id": VARCHAR2(4000) (REQ)}, \texttt{"event\_type": VARCHAR2(4000)}, \texttt{"timestamp": TIMESTAMP}, \texttt{"details": VARCHAR2(4000)}
    \item \textbf{"RiskAssessment"}: \texttt{"assessment\_id": VARCHAR2(4000) (REQ)}, \texttt{"score": NUMBER(10)}, \texttt{"assessed\_at": TIMESTAMP}, \texttt{"risk\_level": VARCHAR2(4000)}
\end{itemize}
\smallskip
\textbf{Edge types (all no properties):} "Initiates", "FundsFrom", "ReceivesTo", "AuthorizedBy", "ProcessedFor", "GovernedBy", "HasAuditLog", "HasRiskAssessment"

\smallskip
\textbf{Relationships:} \\
\texttt{(IS "Customer")-[IS "Initiates"]->(IS "PaymentTransaction")} \\
\texttt{(IS "PaymentTransaction")-[IS "FundsFrom"]->(IS "Account")} \\
\texttt{(IS "PaymentTransaction")-[IS "ReceivesTo"]->(IS "Account")} \\
\texttt{(IS "PaymentTransaction")-[IS "AuthorizedBy"]->(IS "PaymentMethod")} \\
\texttt{(IS "PaymentTransaction")-[IS "ProcessedFor"]->(IS "Merchant")} \\
\texttt{(IS "PaymentTransaction")-[IS "GovernedBy"]->(IS "ComplianceRule")} \\
\texttt{(IS "PaymentTransaction")-[IS "HasAuditLog"]->(IS "AuditLog")} \\
\texttt{(IS "PaymentTransaction")-[IS "HasRiskAssessment"]->(IS "RiskAssessment")}
\end{tcolorbox}

\clearpage
\subsection{Zero Shot Prompt}
In \Cref{tab:cypher_zero_prompt,tab:gql_zero_prompt,tab:sql_pgq_zero_prompt}, we show the prompt for zero shot evaluation of openCypher, ISO GQL, and ISO SQL/PGQ respectively.

\promptcaption{Zero shot Prompt for Cypher generation.}{tab:cypher_zero_prompt}
\begin{tcolorbox}[appendixprompt,title=Graph Query Prompt]

\small
You are an expert in graph query languages, specifically openCypher.

\smallskip
\textbf{Schema:} \\
\texttt{\{schema\_text\}}

\smallskip
\textbf{Domain knowledge:} \\
\texttt{\{specific\_knowledge\}}

\smallskip
\textbf{Task:} Convert the user's natural language question into a openCypher query.

\smallskip
\textbf{Output:} Return only the query string.
\end{tcolorbox}

\promptcaption{Zero shot Prompt for GQL generation.}{tab:gql_zero_prompt}
\begin{tcolorbox}[appendixprompt,title=Graph Query Prompt]

\small
You are an expert in graph query languages, specifically ISO GQL (ISO/IEC 39075).

\smallskip
\textbf{Schema (DDL):} \\
\texttt{\{schema\_text\}}

\smallskip
\textbf{Domain knowledge:} \\
\texttt{\{specific\_knowledge\}}

\smallskip
\textbf{Task:} Convert the user's natural language question into a ISO GQL query.

\smallskip
\textbf{Output:} Return only the query string.
\end{tcolorbox}

\promptcaption{Zero shot Prompt for ISO SQL/PGQ generation.}{tab:sql_pgq_zero_prompt}
\begin{tcolorbox}[appendixprompt,title=Graph Query Prompt]

\small
You are an expert in graph query languages, specifically ISO SQL/PGQ (ISO/IEC 9075-16).

\smallskip
\textbf{Schema (DDL):} \\
\texttt{\{schema\_text\}}

\smallskip
\textbf{Property graph name:} \\
\texttt{\{graph\_name\}}

\smallskip
\textbf{Domain knowledge:} \\
\texttt{\{specific\_knowledge\}}

\smallskip
\textbf{Task:} Convert the user's natural language question into an ISO SQL/PGQ query.

\smallskip
\textbf{Output:} Return only the query string.
\end{tcolorbox}

\clearpage
\subsection{Few Shot Prompt}
In \Cref{tab:cypher_few_prompt,tab:gql_few_prompt,tab:sql_pgq_few_prompt}, we show the prompt for few shot evaluation of openCypher, ISO GQL, and ISO SQL/PGQ respectively. Compared to the zero-shot prompts, the few-shot prompts additionally include three question--query examples to guide the model.

\promptcaption{Few shot Prompt for Cypher generation.}{tab:cypher_few_prompt}
\begin{tcolorbox}[appendixprompt,title=Graph Query Prompt]

\small
You are an expert in graph query languages, specifically openCypher.

\smallskip
\textbf{The database schema is as follows:} \\
\texttt{\{schema\_text\}}

\smallskip
\textbf{Domain Knowledge:} \\
\texttt{\{specific\_knowledge\}}

\smallskip
\textbf{Task:} Convert the user's natural language question into a openCypher query.

\smallskip
\textbf{Output:} Return only the query string.

\smallskip
\textbf{The user's question and corresponding output examples are as follows:}

\smallskip
\textbf{Example 1} \\
\textbf{Question:} Which characters have a path to ``Catelyn-Stark'' in the interaction network with a maximum of 3 hops? \\
\textbf{Output:} \texttt{MATCH (c:Character)-[:INTERACTS*1..3]->(target:Character \{name: 'Catelyn-Stark'\}) RETURN DISTINCT c.name}

\smallskip
\textbf{Example 2} \\
\textbf{Question:} How many people have directed more than two movies? \\
\textbf{Output:} \texttt{MATCH (p:Person)-[:DIRECTED]->(m:Movie) WITH p, count(m) AS moviesDirected WHERE moviesDirected > 2 RETURN count(p) AS directorsCount}

\smallskip
\textbf{Example 3} \\
\textbf{Question:} List the top 5 movies with the most production companies involved. \\
\textbf{Output:} \texttt{MATCH (m:Movie)-[:PRODUCED\_BY]->(pc:ProductionCompany) WITH m, COUNT(pc) AS productionCompanyCount ORDER BY productionCompanyCount DESC LIMIT 5 RETURN m.title AS MovieTitle, productionCompanyCount}
\end{tcolorbox}

\promptcaption{Few shot Prompt for GQL generation.}{tab:gql_few_prompt}
\begin{tcolorbox}[appendixprompt,title=Graph Query Prompt]

\small
You are an expert in graph query languages, specifically ISO GQL (ISO/IEC 39075).

\smallskip
\textbf{The database schema is as follows:} \\
\texttt{\{schema\_text\}}

\smallskip
\textbf{Domain Knowledge:} \\
\texttt{\{specific\_knowledge\}}

\smallskip
\textbf{Task:} Convert the user's natural language question into a ISO GQL query.

\smallskip
\textbf{Output:} Return only the query string.

\smallskip
\textbf{The user's question and corresponding output examples are as follows:}

\smallskip
\textbf{Example 1} \\
\textbf{Question:} Which characters have a path to ``Catelyn-Stark'' in the interaction network with a maximum of 3 hops? \\
\textbf{Output:} \texttt{MATCH (c:Character)-[:INTERACTS]->\{1,3\}(target:Character \{name: 'Catelyn-Stark'\}) RETURN DISTINCT c.name}

\smallskip
\textbf{Example 2} \\
\textbf{Question:} How many people have directed more than two movies? \\
\textbf{Output:} \texttt{MATCH (p:Person)-[:DIRECTED]->(m:Movie) RETURN p, count(m) AS moviesDirected NEXT FILTER moviesDirected > 2 RETURN count(p) AS directorsCount}

\smallskip
\textbf{Example 3} \\
\textbf{Question:} List the top 5 movies with the most production companies involved. \\
\textbf{Output:} \texttt{MATCH (m:Movie)-[:PRODUCED\_BY]->(pc:ProductionCompany) RETURN m, COUNT(pc) AS productionCompanyCount ORDER BY productionCompanyCount DESC LIMIT 5 NEXT RETURN m.title AS MovieTitle, productionCompanyCount}
\end{tcolorbox}

\clearpage
\promptcaption{Few shot Prompt for ISO SQL/PGQ generation.}{tab:sql_pgq_few_prompt}
\begin{tcolorbox}[appendixprompt,title=Graph Query Prompt]

\small
You are an expert in graph query languages, specifically ISO SQL/PGQ (ISO/IEC 9075-16).

\smallskip
\textbf{The database schema is as follows:} \\
\texttt{\{schema\_text\}}

\smallskip
\textbf{Property graph name:} \\
\texttt{\{graph\_name\}}

\smallskip
\textbf{Domain Knowledge:} \\
\texttt{\{specific\_knowledge\}}

\smallskip
\textbf{Task:} Convert the user's natural language question into an ISO SQL/PGQ query.

\smallskip
\textbf{Output:} Return only the query string.

\smallskip
\textbf{The user's question and corresponding output examples are as follows:}

\smallskip
\textbf{Example 1} \\
\textbf{Question:} Which characters have a path to ``Catelyn-Stark'' in the interaction network with a maximum of 3 hops? \\
\textbf{Output:} \texttt{SELECT DISTINCT * FROM GRAPH\_TABLE (} \\
\texttt{~~"\{graph\_name\}" MATCH (c IS "Character")-[e1 IS "INTERACTS"]->\{1,3\}(target IS "Character") WHERE target."name" = 'Catelyn-Stark' COLUMNS (c."name" AS name)} \\
\texttt{) gt}

\smallskip
\textbf{Example 2} \\
\textbf{Question:} How many people have directed more than two movies? \\
\textbf{Output:} \texttt{WITH stage\_1 AS (} \\
\texttt{~~SELECT p\_VALUE, COUNT(m\_VALUE) AS moviesDirected} \\
\texttt{~~FROM GRAPH\_TABLE (} \\
\texttt{~~~~"\{graph\_name\}" MATCH (p IS "Person")-[e1 IS "DIRECTED"]->(m IS "Movie") COLUMNS (VERTEX\_ID(p) AS p\_VALUE, VERTEX\_ID(m) AS m\_VALUE)} \\
\texttt{~~) gt} \\
\texttt{~~GROUP BY p\_VALUE} \\
\texttt{)} \\
\texttt{SELECT COUNT(p\_VALUE) AS directorsCount} \\
\texttt{FROM stage\_1} \\
\texttt{WHERE moviesDirected > 2}

\smallskip
\textbf{Example 3} \\
\textbf{Question:} List the top 5 movies with the most production companies involved. \\
\textbf{Output:} \texttt{WITH stage\_1 AS (} \\
\texttt{~~SELECT m\_VALUE, COUNT(pc\_VALUE) AS productionCompanyCount, m\_title} \\
\texttt{~~FROM GRAPH\_TABLE (} \\
\texttt{~~~~"\{graph\_name\}" MATCH (m IS "Movie")-[e1 IS "PRODUCED\_BY"]->(pc IS "ProductionCompany") COLUMNS (VERTEX\_ID(m) AS m\_VALUE, VERTEX\_ID(pc) AS pc\_VALUE, m."title" AS m\_title)} \\
\texttt{~~) gt} \\
\texttt{~~GROUP BY m\_VALUE, m\_title} \\
\texttt{~~ORDER BY productionCompanyCount DESC} \\
\texttt{~~FETCH FIRST 5 ROWS ONLY} \\
\texttt{)} \\
\texttt{SELECT m\_title AS MovieTitle, productionCompanyCount AS productionCompanyCount} \\
\texttt{FROM stage\_1}
\end{tcolorbox}

\clearpage
\subsection{Synthesis Prompt}
The data synthesis pipeline consists of four stages executed in sequence: (1)~\textit{Schema Description Generation}---designing a graph schema for a specified domain with business context (\Cref{tab:schema_desc_gen_prompt}); (2)~\textit{Schema Generation}---converting the schema description into a formatted, machine-readable schema (\Cref{tab:schema_gen_prompt}); (3)~\textit{Data Generation}---creating realistic CSV data via a Python script (\Cref{tab:data_gen_prompt}); and (4)~\textit{Corpus Generation}---synthesizing executable question--query pairs (\Cref{tab:corpus_gen_prompt}). Below we present the key prompts for each stage.

\promptcaption{Schema Description Generator Prompt.}{tab:schema_desc_gen_prompt}
\begin{tcolorbox}[appendixprompt,title=Schema Description Generator Prompt]
\lstset{basicstyle=\small, escapechar=|}

\promptsectiontitle{System Prompt}
\begin{lstlisting}
You are a business domain modeling specialist with expertise in graph-based knowledge representation. Your task is to create detailed documentation for graph schemas that accurately model real-world subdomains.

Structure your response with these sections:
1. Subdomain Introduction
2. Core Node Types (Entities)
3. Relationship Types (Connections)
4. Schema Diagram (using Mermaid syntax)
5. Business Rules & Constraints
6. Example Usage Scenarios

|\textbf{Instruction:}|
Generate a complete graph schema documentation for:
Domain: {domain}
Subdomain: {subdomain}
Key Business Requirements:
1. [Critical business objective #1]
2. [Important operational constraint #2]
3. [Key data interaction pattern #3]

Specific Context Details:
- Industry: [e.g., Healthcare, FinTech]
- Scale: [e.g., 100K daily transactions]
- Critical Relationships: [e.g., Regulatory dependencies]
- Special Constraints: [e.g., Privacy compliance needs]

Output Requirements:
- Use business-friendly terminology
- Include 4-6 core node types
- Define 3-5 relationship types with clear semantics
- Provide Mermaid diagram for visualization
- Explain how schema supports business requirements
\end{lstlisting}
\end{tcolorbox}

\clearpage
\promptcaption{Schema Generator Prompt.}{tab:schema_gen_prompt}
\begin{tcolorbox}[appendixprompt,title=Schema Generator Prompt]
\lstset{basicstyle=\small, escapechar=|}

\promptsectiontitle{System Prompt}
\begin{lstlisting}
You are an expert graph database architect with 15+ years of experience designing schemas for complex domains.
Your specialty is creating performant, intuitive graph models that balance normalization with real-world query needs.
Design a comprehensive graph schema for my target domain.
Follow this thinking framework:
|\textbf{Instruction:}|
Schema Description:
{schema_description}

You are a top-tier graph database architect. Please design a professional Schema for the {domain} domain and {subdomain} subdomain base on the Schema Description.

Critical Format Requirements:
1. For VERTEX nodes:
   - Every VERTEX MUST have a dedicated ID property named using the format: `[LABEL]_id` (case-sensitive)
   - "primary" MUST be defined at the SAME LEVEL as "properties", NOT inside properties array
2. For EDGE relationships:
   - Only ONE entry per relationship type
   - Multiple constraints MUST be consolidated into SINGLE "constraints" array

|\textbf{Task Requirements}|
1. Include {min_nodes}-{max_nodes} node types
2. Include {min_rels}-{max_rels} relationship types
3. Strict schema format compliance:
    schema (array)
    label (required, string)
    type (required: only VERTEX/EDGE)
    properties (array, required for vertices)
      name (required, string)
      type (required: BOOL,INT8,INT16,INT32,INT64,
      DATE,DATETIME,FLOAT,DOUBLE,STRING,BLOB)
      optional / index / unique / pair_unique (optional)
    primary (VERTEX ONLY, REQUIRED, TOP-LEVEL FIELD)
    temporal / temporal_field_order (EDGE ONLY, optional)
    constraints (EDGE ONLY, optional, array of label pairs)
    detach_property (optional, default false)

|\textbf{Prohibited Patterns}|
NEVER include "primary" inside properties array
NEVER create multiple entries for same edge type
NEVER combine unique/pair_unique in same property

|\textbf{Output Example}|
{example_json}

|\textbf{Compliance Verification}|
Before final output, self-check:
1. Every VERTEX has top-level "primary" (single property name, no comma-separated values)
2. Every EDGE has SINGLE entry with consolidated constraints
3. ZERO "primary" in properties arrays
4. Naming: UPPER_SNAKE_CASE for node labels, PascalCase for relationship types
5. Case consistency across all schema element references
\end{lstlisting}
\end{tcolorbox}

\clearpage
\promptcaption{Data Generator Prompt.}{tab:data_gen_prompt}
\begin{tcolorbox}[appendixprompt,title=Data Generator Prompt]
\lstset{basicstyle=\small, escapechar=|}

\promptsectiontitle{System Prompt}
\begin{lstlisting}
You are a world-class Python engineer and an expert in graph database test data generation.
Your task is to generate a single, self-contained, and executable Python script based on a given graph schema.
The script must produce high-quality, realistic data that simulates complex real-world scenarios.

|\textbf{Instruction:}|
Graph Schema:
```json
{schema_json}
```
As a top-tier Python engineer, create a professional, high-quality Python script to generate realistic test data based on the provided graph schema.

Core Requirements:
0. Strict Schema Adherence (TOP PRIORITY): CSV headers and columns MUST STRICTLY match the properties defined in the schema_json. No extra columns (e.g., `id`, `_id`).
1. Single Script: Directly executable (`python your_script.py`) without modifications.
2. Libraries: Only `csv`, `faker`, `numpy`, `random`, `datetime`, `os`. Import Faker via `from faker import Faker`.
3. Reproducibility: `Faker('en_US')`, `np.random.seed(42)`, `random.seed(42)`.
4. File Output: One CSV per node/edge type, saved to `./csv_files/` relative to the script.
5. OOP Script Generation: Use a class-based structure.

Data Realism & Quality:
6. Data Volume: Configurable parameters (e.g., `USER_COUNT`, `TRANSACTION_COUNT`).
7. Realistic Distributions: power-law (zipf) for hub nodes, log-normal for amounts, normal for age/scores.
8. Data Consistency: Node-generation functions MUST return unique ID lists; IDs passed as function arguments (no global variables).
9. Edge CSV Format: Headers use source/destination node type names (not `_from`/`_to`); values are raw node IDs only.
10. Data Noise: ~10

CSV Formatting:
11. `csv.writer` with `newline=''`, `encoding='utf-8'`, `quoting=csv.QUOTE_MINIMAL`.
12. Sanitize all rows before writing: format datetime to `YYYY-MM-DD HH:MM:SS`, convert `None` to `''`, strip whitespace/newlines.
13. Every row must have the exact same number of columns as the header.

Output: Python code only, self-contained, with `if __name__ == "__main__":` block.
\end{lstlisting}
\end{tcolorbox}

\clearpage
\promptcaption{Corpus Generator Prompt.}{tab:corpus_gen_prompt}
\begin{tcolorbox}[appendixprompt,title=Corpus Generator Prompt]
\lstset{basicstyle=\small, escapechar=@}

\promptsectiontitle{System Prompt}
\begin{lstlisting}
You are an expert in graph databases and the Cypher query language. Your task is to generate new, high-quality, and diverse "natural language question-Cypher query" data pairs based on the provided graph schema and some validated query examples.
Your output must be in strict JSON format, use English, as a list containing multiple objects.

@\textbf{Instruction:}@
Generate {num_per_iteration} new "question-query" data pairs based on the following information.

1. Graph Schema: {schema_json}
2. Verified Query Examples: {examples_json}

3. Your Task: Generate {num_per_iteration} new, more interesting, and potentially more complex "question-query" data pairs.
Guidelines:
- Diversity: aggregations (COUNT, SUM, AVG), filtering (WHERE), multi-hop, optional matching, etc.
- Increasing Complexity: more complex than examples, but logically meaningful.
- No Repetition: Do not duplicate existing examples.
- Critical: NEVER return a whole node/relationship; always select specific properties.
- Output: JSON list of {{"question": "...", "query": "..."}} objects.

@\textbf{Enhancement:}@
Create more complex "question-query" pairs by:
- Combining query patterns from multiple examples
- Extending path queries with more hops
- Using advanced functions (COUNT, SUM, AVG, COLLECT, OR, NOT, IN)
- Asking deeper analytical questions ("why", "compare", etc.)

@\textbf{Query Archetypes (8 categories):}@
1. Aggregation and Counting
2. Filtering and Sorting
3. Relationship Reachability
4. Multi-hop Path Query
5. Common Neighbors and Association Analysis
6. Existence and Boolean Checks
7. Attribute Comparison Query
8. Path Analysis and Traversal

@\textbf{Query Templates (selected):}@
Basic: MATCH (n:label) WHERE n.prop = "val" RETURN n.prop LIMIT 10
One-hop: MATCH (a:label_1)-[:edge]->(b:label_2) WHERE b.status = "active" RETURN a.id
Variable-length: MATCH (n1:label_1)-[*1..3]->(n2:label_2) WHERE n1.prop = "val" RETURN n2.id LIMIT 5
Aggregation: MATCH (n:label) WHERE n.score > 60 RETURN n.name, n.score ORDER BY n.score DESC LIMIT 5
Shortest path: MATCH p = shortestPath((a:label_1)-[:edge*1..10]->(b:label_3)) WHERE a.prop = "val" RETURN p, length(p) AS depth LIMIT 5
Optional match: MATCH (a:label_1)-[:edge_1]->(b:label_2) WHERE a.prop = "val" OPTIONAL MATCH (b)<-[:edge_2]-(c:label_3) RETURN c.prop LIMIT 10
Multi-step (WITH): MATCH (n:label_1) WHERE n.prop = "val" WITH n MATCH (n)-[:edge]->(m:label_2) RETURN m LIMIT 10

@\textbf{Exploration Prompt:}@
Brainstorm diverse natural language questions around a given "query intent" archetype, using the schema and verified examples as context. Output: JSON list of question strings.

@\textbf{Translation Prompt:}@
Translate a natural language question into a Cypher query.
Rules:
1. Attributes must belong to the correct node/edge in the schema.
2. Never use edge attributes on nodes (access via relationship variable).
3. Use only schema-defined elements.
4. Use '
Output: JSON object with single "query" key.
\end{lstlisting}
\end{tcolorbox}

\clearpage
\subsection{Annotation Prompt}

As described in Section~\ref{subsec:question_aliasing}, each query is annotated at three hierarchical abstraction levels: \textbf{Syntactic Level} (explicit graph query primitives), \textbf{Logical Level} (entity-relation logic without keywords), and \textbf{Business Level} (high-level analytical intent) (\Cref{tab:nl_desc_prompt}). For Business Level questions, external knowledge is derived by reverse-rationalizing aligned question--query pairs to make implicit schema or domain assumptions explicit (\Cref{tab:ext_know_prompt}). Additionally, schema element names in questions are replaced with semantically similar aliases; an alias is accepted only if its embedding's closest match in the full schema remains the original entity (\Cref{tab:schema_aliasing_prompt}). Below we present the annotation prompts for three-level question generation, external knowledge derivation, and schema aliasing.

\promptcaption{Annotation Prompt for 3-Level Questions Descriptions.}{tab:nl_desc_prompt}
\begin{tcolorbox}[appendixprompt,title=Three-Level Question Generation Prompt]

\small
You are an expert in Graph Query Language (GQL) and natural language understanding. Your task is to generate 3 levels of natural language descriptions for the given GQL query, following the framework from ``Accessible Visualization via Natural Language Descriptions: A Three-Level Model of Semantic Content''.

\smallskip
\promptsectiontitle{3-Level Framework Definition}

\smallskip
\textbf{Level 1: Structural/Syntactic Elements}
\begin{itemize}
    \setlength\itemsep{0em}
    \item Purpose: Basic query components and syntax
    \item Characteristics: Directly mentions node labels, relationship types, properties, and graph patterns
    \item Example: ``List `Entity' that were incorporated on `23-MAR-2006'.''
\end{itemize}

\smallskip
\textbf{Level 2: Semantic/Logical Operations}
\begin{itemize}
    \setlength\itemsep{0em}
    \item Purpose: Query logic and computational meaning
    \item Characteristics: Describes what the query does without explicit graph terminology, focuses on the logical operation
    \item Example: ``List entities incorporated in March 23, 2006.''
\end{itemize}

\smallskip
\textbf{Level 3: Analytical/Business Patterns}
\begin{itemize}
    \setlength\itemsep{0em}
    \item Purpose: Data analysis strategy and methodological insights
    \item Characteristics: Describes the analytical approach, patterns being investigated, or research methodology
    \item Example: ``Analyze company incorporation trends over 2026.''
\end{itemize}

\smallskip
\noindent\rule{\linewidth}{0.4pt}

\smallskip
\promptsectiontitle{Given Information}

\smallskip
\textbf{GQL Query:} \\
\texttt{\{gql\}}

\smallskip
\textbf{Schema Context:} \\
\texttt{\{db\_schema\}}

\smallskip
\noindent\rule{\linewidth}{0.4pt}

\smallskip
\promptsectiontitle{Task}

Based on the GQL query above, generate 3 levels of natural language descriptions.

\smallskip
\promptsectiontitle{Important Guidelines}
\begin{enumerate}
    \setlength\itemsep{0em}
    \item Each level should be a complete, standalone natural language question or statement
    \item Level 1 should include explicit graph terminology (nodes, relationships, properties)
    \item Level 2 should be semantically clear without graph-specific terms
    \item Level 3 should focus on analytical methodology or data exploration strategy
    \item Ensure smooth progression from concrete (L1) to abstract (L3)
    \item Each level should be different from the others in abstraction and focus
\end{enumerate}

\smallskip
\promptsectiontitle{Output Format}
\texttt{\{output\_example\}}

\end{tcolorbox}

\clearpage
\promptcaption{Annotation Prompt for External Knowledge.}{tab:ext_know_prompt}
\begin{tcolorbox}[appendixprompt,title=External Knowledge Prompt]

\small
You are an expert in Graph Query Language (GQL) and natural language (NL) understanding. Your task is to generate the external knowledge for high level NL based on the gql query and basic NL query. The external knowledge should be specific and clear, and should be able to help the model understand the high level NL query better.

\smallskip
\noindent\rule{\linewidth}{0.4pt}

\smallskip
\promptsectiontitle{Given Information}

\smallskip
\textbf{GQL Query:} \\
\texttt{\{gql\}}

\smallskip
\textbf{Basic NL:} \\
\texttt{\{level\_2\}}

\smallskip
\textbf{High Level NL:} \\
\texttt{\{level\_3\}}

\smallskip
\noindent\rule{\linewidth}{0.4pt}

\smallskip
\promptsectiontitle{Task}
Generate external knowledge that bridges the gap between the High Level NL and Basic NL queries.

\smallskip
\promptsectiontitle{Guidelines}
\begin{enumerate}
    \setlength\itemsep{0em}
    \item \textbf{Identify missing specifics:} Extract exact values, thresholds, entity names, or technical terms from the GQL/Basic NL that are absent or vague in the High Level NL.
    \item \textbf{Be precise:} State concrete numbers, definitions, or clarifications (e.g., ``a high rating means rating $>=$ 4.0'', ``recent refers to the last 30 days'').
    \item \textbf{Keep it minimal:} Limit to 1-2 sentences (max 50 words). Only include what is necessary to disambiguate the High Level NL.
    \item \textbf{Return empty if unnecessary:} If the High Level NL already contains all specifics from the Basic NL, return an empty string ``''.
\end{enumerate}

\smallskip
\promptsectiontitle{Output Format}
\texttt{\{output\_example\}}

\end{tcolorbox}

\clearpage

\promptcaption{Schema-Level Alias Candidate Generation Prompt.}{tab:stage1_alias_prompt}
\begin{tcolorbox}[appendixprompt,title=Stage 1 Schema Alias Candidate Generation Prompt]

\small
You are a precise NLP assistant for Graph Query Language (GQL) data augmentation.

You are given a graph schema. For EACH entity (vertex label, relationship type, or property), you are also given its \texttt{plain} form (the natural-text reading of the identifier). Your job is to produce multiple \texttt{alias\_candidates}: SYNONYMS of \texttt{plain} made of DIFFERENT WORDS that carry the SAME meaning, suitable for substituting into English sentences about the schema.

\smallskip
\promptsectiontitle{Hard Requirements}
\begin{enumerate}
    \setlength\itemsep{0em}
    \item For compound / domain-specific identifiers (CamelCase labels, snake\_case composites, multi-word phrases, identifiers with a domain-specific noun), the alias MUST be different lexical content from \texttt{plain}. A mere surface form variant (lowercase, plural, singular, snake-case-to-space) is FORBIDDEN in this case.
    \item \textbf{EXCEPTION:} if \texttt{plain} is ALREADY the most natural single English word for the concept and any synonym would either (a) feel awkward in normal English, (b) change the meaning, or (c) introduce ambiguity with another entity in this schema, then KEEP \texttt{alias == plain}. This is the preferred behaviour for bare common nouns like ``name'', ``amount'', ``status'', ``description'', ``details'', ``currency'', ``phone'', ``email'', ``address'', ``score'', and ``balance''. Forcing a synonym onto these words usually distorts meaning (e.g., ``name'' $\rightarrow$ ``label'' is WRONG because ``label'' means something different in a graph database).
    \item \textbf{Meaning preserved:} the alias must mean the SAME thing as \texttt{plain} for THIS entity in THIS schema. Do not generalise, specialise, or invent fields.
    \item \textbf{No cross-entity ambiguity:} candidates for entity A MUST NOT be more natural descriptions of any OTHER entity B in the same schema. Read the WHOLE input first; consider every other label / edge / property before committing to candidates. If two entities are semantically close (e.g., several date-like or status-like properties), propose candidates that uniquely identify each one.
    \item Aliases for labels / properties: short noun phrases. Aliases for relationship types: short verb phrases.
    \item Return 2--4 candidates per entity when possible. Include \texttt{plain} as a safe candidate when no synonym is clearly safe.
    \item Return STRICT JSON ONLY (no markdown, no code fences).
\end{enumerate}

\smallskip
\promptsectiontitle{Good vs. Bad Aliases}
\begin{itemize}
    \setlength\itemsep{0em}
    \item \texttt{Customer} (plain: ``customer'') $\rightarrow$ GOOD: ``client''; BAD: ``customer'', ``customers'' (mere normalisation)
    \item \texttt{PaymentTransaction} (plain: ``payment transaction'') $\rightarrow$ GOOD: ``money transfer''; BAD: ``payment transaction'', ``payment record'' (only normalised)
    \item \texttt{AuthorizedBy} (plain: ``authorized by'') $\rightarrow$ GOOD: ``approved by''; BAD: ``authorized by''
    \item \texttt{business\_name} (plain: ``business name'') $\rightarrow$ GOOD: ``brand name''; BAD: ``business name''
    \item \texttt{name} (plain: ``name'') $\rightarrow$ GOOD: ``name''; BAD: ``label'' (changes meaning in a graph DB)
    \item \texttt{amount} (plain: ``amount'') $\rightarrow$ GOOD: ``amount''; BAD: ``total'' (more specific than amount)
    \item \texttt{status} (plain: ``status'') $\rightarrow$ GOOD: ``status''; BAD: ``condition'' (changes meaning)
    \item \texttt{description} (plain: ``description'') $\rightarrow$ GOOD: ``description''; BAD: ``summary'' (specific subtype)
    \item \texttt{score} (plain: ``score'') $\rightarrow$ GOOD: ``score''; BAD: ``rating'' if rating means something different here
\end{itemize}

\smallskip
\promptsectiontitle{Cross-Entity Ambiguity}
A schema may have several semantically-adjacent fields. Choose aliases so the mapping back from alias to entity stays UNIQUE. Concrete example:

\smallskip
Suppose the schema has BOTH:
\begin{itemize}
    \setlength\itemsep{0em}
    \item \texttt{Account.opened\_date} (plain: ``opened date'')
    \item \texttt{ComplianceRule.effective\_date} (plain: ``effective date'')
\end{itemize}

\smallskip
\textbf{BAD aliasing (the words get SWAPPED in meaning):}
\begin{itemize}
    \setlength\itemsep{0em}
    \item \texttt{Account.opened\_date} $\rightarrow$ ``activation date'' (sounds more like a rule going into effect)
    \item \texttt{ComplianceRule.effective\_date} $\rightarrow$ ``start date'' (sounds more like an account opening)
\end{itemize}

\smallskip
\textbf{GOOD aliasing (each alias points uniquely back at its own entity):}
\begin{itemize}
    \setlength\itemsep{0em}
    \item \texttt{Account.opened\_date} $\rightarrow$ ``creation date'' or ``establishment date''
    \item \texttt{ComplianceRule.effective\_date} $\rightarrow$ ``in-force date'' or ``enforcement date''
\end{itemize}

Before returning, do a quick self-check: for every candidate, ask ``could this alias plausibly describe a DIFFERENT entity in the same schema better than the one I assigned it to?'' If yes, pick a different candidate or include \texttt{plain}.

\smallskip
\noindent\rule{\linewidth}{0.4pt}

\smallskip
\promptsectiontitle{Given Information}

\smallskip
\textbf{Input Schema JSON:} \\
\texttt{\{stage1\_input\}}

\smallskip
The input JSON contains the domain and all schema entities in the following groups: \texttt{labels}, \texttt{edges}, and \texttt{properties}. Each entity contains its schema \texttt{name} and deterministic \texttt{plain} form.

\smallskip
\noindent\rule{\linewidth}{0.4pt}

\smallskip
\promptsectiontitle{Task}
Produce the alias map for this schema and return STRICT JSON only.

\smallskip
\promptsectiontitle{Output Format}
Return an object with exactly the following schema:

\smallskip
\texttt{\{}\\
\hspace*{1em}\texttt{"labels": \{"<Label>": ["<candidate1>", "<candidate2>", ...], ...\},}\\
\hspace*{1em}\texttt{"edges": \{"<EdgeType>": ["<candidate1>", "<candidate2>", ...], ...\},}\\
\hspace*{1em}\texttt{"properties": \{"<Label>.<prop>": ["<candidate1>", "<candidate2>", ...], ...\}}\\
\texttt{\}}

\smallskip
Every key from the input MUST appear in the output. Do not add extra keys.

\end{tcolorbox}

\clearpage
\promptcaption{Per-Sample Question Rewrite Prompt Using Schema Aliases.}{tab:stage2_rewrite_prompt}
\begin{tcolorbox}[appendixprompt,title=Stage 2 Per-Sample Question Rewrite Prompt]

\small
You are a precise NLP assistant for Graph Query Language (GQL) data augmentation.

\smallskip
\promptsectiontitle{Given Information}
You are given:
\begin{itemize}
    \setlength\itemsep{0em}
    \item \texttt{initial\_question} and \texttt{initial\_gql};
    \item \texttt{extracted\_schema\_entities\_from\_gql}: deterministic symbols parsed from this query (labels, edges, resolved/unresolved property refs, \texttt{RETURN ... AS} names). Treat this as the list of schema-related tokens present in the query text;
    \item \texttt{candidate\_aliases}: ONLY the subset of the domain schema map that matches those extracted keys. Each entry has \texttt{plain} and \texttt{alias}. You MUST NOT invent aliases or pull synonyms from outside \texttt{candidate\_aliases}.
\end{itemize}

\smallskip
\textbf{Input JSON:} \\
\texttt{\{stage2\_input\}}

\smallskip
\noindent\rule{\linewidth}{0.4pt}

\smallskip
\promptsectiontitle{Task}
\begin{enumerate}
    \setlength\itemsep{0em}
    \item From \texttt{extracted\_schema\_entities\_from\_gql}, decide which tokens actually have a matching mention in \texttt{initial\_question} (match loosely: case, plurals, mild inflection). Skip tokens with no good NL anchor unless \texttt{RETURN}/order wording clearly aligns (e.g., schema token \texttt{total\_amount} $\leftrightarrow$ phrase ``total amount''). Tokens that appear only in the query with no NL counterpart may be omitted from entity/alias rows.
    \item For each row you keep, look up \texttt{plain} and \texttt{alias} in \texttt{candidate\_aliases} when the token is a mapped label, edge, or \texttt{"Label.prop"} property key; use identity (repeat the NL mention in \texttt{alias[i][1]}) when you choose not to substitute or when there is no entry for that token.
    \item Produce \texttt{new\_initial\_question}: substitute chosen mentions with \texttt{alias} only where fluent and meaning-preserving.
\end{enumerate}

\smallskip
\promptsectiontitle{Output Rules}
\begin{itemize}
    \setlength\itemsep{0em}
    \item \texttt{entity[i] = ["<schema\_token\_from\_gql>", "<matched\_text\_in\_initial\_question>"]} --- same breadth intent as standalone \texttt{entity\_replace} (full alignment trace).
    \item \texttt{alias} MUST have the SAME LENGTH and SAME ORDER as \texttt{entity}; \texttt{alias[i][0] == entity[i][0]}; \texttt{alias[i][1]} is the surface form in \texttt{new\_initial\_question} (mapped \texttt{alias} when substituted, otherwise repeat \texttt{entity[i][1]}).
    \item Property rows MUST use the same \texttt{"<Label>.<prop>"} keys as in \texttt{candidate\_aliases} when referring to node properties.
    \item Do NOT modify \texttt{initial\_gql}.
    \item Return STRICT JSON ONLY (no markdown, no code fences, no commentary).
\end{itemize}

\smallskip
\promptsectiontitle{Output Format}
Return an object with exactly the following schema:

\smallskip
\texttt{\{}\\
\hspace*{1em}\texttt{"id": "...",}\\
\hspace*{1em}\texttt{"initial\_question": "...",}\\
\hspace*{1em}\texttt{"initial\_gql": "...",}\\
\hspace*{1em}\texttt{"entity": [["...", "..."]],}\\
\hspace*{1em}\texttt{"alias": [["...", "..."]],}\\
\hspace*{1em}\texttt{"new\_initial\_question": "..."}\\
\texttt{\}}

\end{tcolorbox}

\clearpage
\subsection{Database Import Config}

As described in Section~\ref{subsec:data_query_gen}, after schemas are finalized, existing datasets are converted through a deterministic pipeline that normalizes formats and generates import configurations for target graph engines. Below we show three examples of database import configurations instantiated from the same \texttt{SchemaGraph} for the Financial Payment domain: a TuGraph-DB import config (\Cref{tab:tugraph_import_config}), a Spanner Graph DDL (\Cref{tab:spanner_import_ddl}), and an Oracle CPG DDL (\Cref{tab:oracle_import_ddl}).

\captionof{table}{TuGraph-DB Import Config}
\label{tab:tugraph_import_config}
\begin{tcolorbox}[
    colback=blue!5!white,
    colframe=blue!75!black,
    title=TuGraph-DB Import Config,
    fonttitle=\bfseries,
    arc=2pt,
    width=\textwidth,
    boxrule=0.5pt,
    left=3pt,
    right=3pt,
    top=3pt,
    bottom=3pt
]

\lstset{
    basicstyle=\scriptsize\ttfamily,
    breaklines=true,
    columns=fullflexible,
    escapechar=|
}

\begin{lstlisting}
{
  "schema": [
    {"type": "VERTEX", "label": "PaymentTransaction",
     "primary": "transaction_id",
     "properties": [
       {"name": "transaction_id", "type": "STRING", "optional": false, "unique": true, "index": true},
       {"name": "amount", "type": "DOUBLE"}, {"name": "currency", "type": "STRING"},
       {"name": "status", "type": "STRING"}, {"name": "created_at", "type": "DATETIME"},
       {"name": "completed_at", "type": "DATETIME"}, {"name": "reference_number", "type": "STRING"}]},
    {"type": "VERTEX", "label": "Account",
     "primary": "account_id",
     "properties": [
       {"name": "account_id", "type": "STRING", "optional": false, "unique": true, "index": true},
       {"name": "account_type", "type": "STRING"}, {"name": "balance", "type": "DOUBLE"},
       {"name": "currency", "type": "STRING"}, {"name": "status", "type": "STRING"},
       {"name": "opened_date", "type": "DATE"}, {"name": "last_activity", "type": "DATETIME"}]},
    {"type": "VERTEX", "label": "Customer",
     "primary": "customer_id",
     "properties": [
       {"name": "customer_id", "type": "STRING", "optional": false, "unique": true, "index": true},
       {"name": "customer_type", "type": "STRING"}, {"name": "name", "type": "STRING"},
       {"name": "email", "type": "STRING"}, {"name": "phone", "type": "STRING"},
       {"name": "address", "type": "STRING"}, {"name": "kyc_status", "type": "STRING"}]},
    {"type": "VERTEX", "label": "PaymentMethod", "primary": "method_id", ...},
    {"type": "VERTEX", "label": "Merchant", "primary": "merchant_id", ...},
    {"type": "VERTEX", "label": "ComplianceRule", "primary": "rule_id", ...},
    {"type": "VERTEX", "label": "AuditLog", "primary": "log_id", ...},
    {"type": "VERTEX", "label": "RiskAssessment", "primary": "assessment_id", ...},
    {"type": "EDGE", "label": "Initiates", "properties": [], "constraints": [["Customer", "PaymentTransaction"]]},
    {"type": "EDGE", "label": "FundsFrom", "properties": [], "constraints": [["PaymentTransaction", "Account"]]},
    {"type": "EDGE", "label": "ReceivesTo", "properties": [], "constraints": [["PaymentTransaction", "Account"]]},
    {"type": "EDGE", "label": "AuthorizedBy", "properties": [], "constraints": [["PaymentTransaction", "PaymentMethod"]]},
    {"type": "EDGE", "label": "ProcessedFor", "properties": [], "constraints": [["PaymentTransaction", "Merchant"]]},
    {"type": "EDGE", "label": "GovernedBy", "properties": [], "constraints": [["PaymentTransaction", "ComplianceRule"]]},
    {"type": "EDGE", "label": "HasAuditLog", "properties": [], "constraints": [["PaymentTransaction", "AuditLog"]]},
    {"type": "EDGE", "label": "HasRiskAssessment", "properties": [], "constraints": [["PaymentTransaction", "RiskAssessment"]]}
  ],
  "files": [
    {"path": "PaymentTransaction.csv", "label": "PaymentTransaction", "format": "CSV", "header": 1,
     "columns": ["transaction_id", "amount", "currency", "status", "created_at", "completed_at", "reference_number"]},
    {"path": "Account.csv", "label": "Account", "format": "CSV", "header": 1,
     "columns": ["account_id", "account_type", "balance", "currency", "status", "opened_date", "last_activity"]},
    {"path": "Customer.csv", "label": "Customer", "format": "CSV", "header": 1,
     "columns": ["customer_id", "customer_type", "name", "email", "phone", "address", "kyc_status"]},
    ...,
    {"path": "Initiates.csv", "label": "Initiates", "format": "CSV", "header": 1,
     "SRC_ID": "Customer", "DST_ID": "PaymentTransaction", "columns": ["SRC_ID", "DST_ID"]},
    {"path": "FundsFrom.csv", "label": "FundsFrom", "format": "CSV", "header": 1,
     "SRC_ID": "PaymentTransaction", "DST_ID": "Account", "columns": ["SRC_ID", "DST_ID"]},
    ...
  ]
}
\end{lstlisting}

\end{tcolorbox}

\clearpage
\captionof{table}{Spanner Graph Import DDL}
\label{tab:spanner_import_ddl}
\begin{tcolorbox}[
    colback=blue!5!white,
    colframe=blue!75!black,
    title=Spanner Graph Import DDL,
    fonttitle=\bfseries,
    arc=2pt, width=\textwidth, boxrule=0.5pt,
    left=3pt, right=3pt, top=3pt, bottom=3pt,
    breakable
]
\lstset{basicstyle=\scriptsize, escapechar=|}
\begin{lstlisting}
CREATE TABLE `PaymentTransaction` (
  `transaction_id` STRING(MAX) NOT NULL,
  `amount` FLOAT64, `currency` STRING(MAX), `status` STRING(MAX),
  `created_at` STRING(MAX), `completed_at` STRING(MAX),
  `reference_number` STRING(MAX)
) PRIMARY KEY (`transaction_id`);

CREATE TABLE `Account` (
  `account_id` STRING(MAX) NOT NULL,
  `account_type` STRING(MAX), `balance` FLOAT64,
  `currency` STRING(MAX), `status` STRING(MAX),
  `opened_date` DATE, `last_activity` STRING(MAX)
) PRIMARY KEY (`account_id`);

CREATE TABLE `Customer` (
  `customer_id` STRING(MAX) NOT NULL,
  `customer_type` STRING(MAX), `name` STRING(MAX),
  `email` STRING(MAX), `phone` STRING(MAX),
  `address` STRING(MAX), `kyc_status` STRING(MAX)
) PRIMARY KEY (`customer_id`);

CREATE TABLE `PaymentMethod` (...) PRIMARY KEY (`method_id`);
CREATE TABLE `Merchant` (...) PRIMARY KEY (`merchant_id`);
CREATE TABLE `ComplianceRule` (...) PRIMARY KEY (`rule_id`);
CREATE TABLE `AuditLog` (...) PRIMARY KEY (`log_id`);
CREATE TABLE `RiskAssessment` (...) PRIMARY KEY (`assessment_id`);

CREATE TABLE `CustomerInitiatesPaymentTransaction` (
  `SRC_ID` STRING(MAX) NOT NULL, `DST_ID` STRING(MAX) NOT NULL,
  FOREIGN KEY (`SRC_ID`) REFERENCES `Customer`(`customer_id`),
  FOREIGN KEY (`DST_ID`) REFERENCES `PaymentTransaction`(`transaction_id`)
) PRIMARY KEY (`SRC_ID`, `DST_ID`);

CREATE TABLE `PaymentTransactionFundsFromAccount` (
  `SRC_ID` STRING(MAX) NOT NULL, `DST_ID` STRING(MAX) NOT NULL,
  FOREIGN KEY (`SRC_ID`) REFERENCES `PaymentTransaction`(`transaction_id`),
  FOREIGN KEY (`DST_ID`) REFERENCES `Account`(`account_id`)
) PRIMARY KEY (`SRC_ID`, `DST_ID`);

CREATE TABLE `PaymentTransactionReceivesToAccount` (...) PRIMARY KEY (`SRC_ID`, `DST_ID`);
CREATE TABLE `PaymentTransactionAuthorizedByPaymentMethod` (...) PRIMARY KEY (`SRC_ID`, `DST_ID`);
CREATE TABLE `PaymentTransactionProcessedForMerchant` (...) PRIMARY KEY (`SRC_ID`, `DST_ID`);
CREATE TABLE `PaymentTransactionGovernedByComplianceRule` (...) PRIMARY KEY (`SRC_ID`, `DST_ID`);
CREATE TABLE `PaymentTransactionHasAuditLogAuditLog` (...) PRIMARY KEY (`SRC_ID`, `DST_ID`);
CREATE TABLE `PaymentTransactionHasRiskAssessmentRiskAssessment` (...) PRIMARY KEY (`SRC_ID`, `DST_ID`);

CREATE OR REPLACE PROPERTY GRAPH `FInancial_Payment`
  NODE TABLES (`PaymentTransaction`, `Account`, `Customer`,
               `PaymentMethod`, `Merchant`, `ComplianceRule`,
               `AuditLog`, `RiskAssessment`)
  EDGE TABLES (
    `CustomerInitiatesPaymentTransaction`
      SOURCE KEY (`SRC_ID`) REFERENCES `Customer`(`customer_id`)
      DESTINATION KEY (`DST_ID`) REFERENCES `PaymentTransaction`(`transaction_id`)
      LABEL `Initiates`,
    `PaymentTransactionFundsFromAccount`
      SOURCE KEY (`SRC_ID`) REFERENCES `PaymentTransaction`(`transaction_id`)
      DESTINATION KEY (`DST_ID`) REFERENCES `Account`(`account_id`)
      LABEL `FundsFrom`,
    `PaymentTransactionReceivesToAccount`
      SOURCE KEY (`SRC_ID`) REFERENCES `PaymentTransaction`(`transaction_id`)
      DESTINATION KEY (`DST_ID`) REFERENCES `Account`(`account_id`)
      LABEL `ReceivesTo`,
    `PaymentTransactionAuthorizedByPaymentMethod`
      SOURCE KEY (`SRC_ID`) REFERENCES `PaymentTransaction`(`transaction_id`)
      DESTINATION KEY (`DST_ID`) REFERENCES `PaymentMethod`(`method_id`)
      LABEL `AuthorizedBy`,
    `PaymentTransactionProcessedForMerchant`
      SOURCE KEY (`SRC_ID`) REFERENCES `PaymentTransaction`(`transaction_id`)
      DESTINATION KEY (`DST_ID`) REFERENCES `Merchant`(`merchant_id`)
      LABEL `ProcessedFor`,
    `PaymentTransactionGovernedByComplianceRule`
      SOURCE KEY (`SRC_ID`) REFERENCES `PaymentTransaction`(`transaction_id`)
      DESTINATION KEY (`DST_ID`) REFERENCES `ComplianceRule`(`rule_id`)
      LABEL `GovernedBy`,
    `PaymentTransactionHasAuditLogAuditLog`
      SOURCE KEY (`SRC_ID`) REFERENCES `PaymentTransaction`(`transaction_id`)
      DESTINATION KEY (`DST_ID`) REFERENCES `AuditLog`(`log_id`)
      LABEL `HasAuditLog`,
    `PaymentTransactionHasRiskAssessmentRiskAssessment`
      SOURCE KEY (`SRC_ID`) REFERENCES `PaymentTransaction`(`transaction_id`)
      DESTINATION KEY (`DST_ID`) REFERENCES `RiskAssessment`(`assessment_id`)
      LABEL `HasRiskAssessment`
  );
\end{lstlisting}
\end{tcolorbox}

\clearpage
\captionof{table}{Oracle CPG Import DDL}
\label{tab:oracle_import_ddl}
\begin{tcolorbox}[
    colback=blue!5!white,
    colframe=blue!75!black,
    title=Oracle CPG Import DDL,
    fonttitle=\bfseries,
    arc=2pt, width=\textwidth, boxrule=0.5pt,
    left=3pt, right=3pt, top=3pt, bottom=3pt,
    breakable
]
\lstset{basicstyle=\scriptsize, escapechar=|}
\begin{lstlisting}
CREATE TABLE "PaymentTransaction_VTX" (
  "transaction_id" VARCHAR2(4000) NOT NULL PRIMARY KEY,
  "amount" FLOAT, "currency" VARCHAR2(4000), "status" VARCHAR2(4000),
  "created_at" TIMESTAMP, "completed_at" TIMESTAMP,
  "reference_number" VARCHAR2(4000)
);
CREATE TABLE "Account_VTX" (
  "account_id" VARCHAR2(4000) NOT NULL PRIMARY KEY,
  "account_type" VARCHAR2(4000), "balance" FLOAT,
  "currency" VARCHAR2(4000), "status" VARCHAR2(4000),
  "opened_date" DATE, "last_activity" TIMESTAMP
);
CREATE TABLE "Customer_VTX" (
  "customer_id" VARCHAR2(4000) NOT NULL PRIMARY KEY,
  "customer_type" VARCHAR2(4000), "name" VARCHAR2(4000),
  "email" VARCHAR2(4000), "phone" VARCHAR2(4000),
  "address" VARCHAR2(4000), "kyc_status" VARCHAR2(4000)
);
CREATE TABLE "PaymentMethod_VTX" (...) PRIMARY KEY ("method_id");
CREATE TABLE "Merchant_VTX" (...) PRIMARY KEY ("merchant_id");
CREATE TABLE "ComplianceRule_VTX" (...) PRIMARY KEY ("rule_id");
CREATE TABLE "AuditLog_VTX" (...) PRIMARY KEY ("log_id");
CREATE TABLE "RiskAssessment_VTX" (...) PRIMARY KEY ("assessment_id");

CREATE TABLE "Initiates_Customer_PaymentTransaction_EDG" (
  "id" NUMBER(19) GENERATED ALWAYS AS IDENTITY PRIMARY KEY,
  "src_Customer_id" VARCHAR2(4000) NOT NULL,
  "dst_PaymentTransaction_id" VARCHAR2(4000) NOT NULL,
  CONSTRAINT "FK_Initiates_SRC" FOREIGN KEY ("src_Customer_id")
    REFERENCES "Customer_VTX"("customer_id"),
  CONSTRAINT "FK_Initiates_DST" FOREIGN KEY ("dst_PaymentTransaction_id")
    REFERENCES "PaymentTransaction_VTX"("transaction_id")
);
CREATE TABLE "FundsFrom_PaymentTransaction_Account_EDG" (
  "id" NUMBER(19) GENERATED ALWAYS AS IDENTITY PRIMARY KEY,
  "src_PaymentTransaction_id" VARCHAR2(4000) NOT NULL,
  "dst_Account_id" VARCHAR2(4000) NOT NULL,
  CONSTRAINT "FK_FundsFrom_SRC" FOREIGN KEY ("src_PaymentTransaction_id")
    REFERENCES "PaymentTransaction_VTX"("transaction_id"),
  CONSTRAINT "FK_FundsFrom_DST" FOREIGN KEY ("dst_Account_id")
    REFERENCES "Account_VTX"("account_id")
);
CREATE TABLE "ReceivesTo_PaymentTransaction_Account_EDG" (...);
CREATE TABLE "AuthorizedBy_PaymentTransaction_PaymentMethod_EDG" (...);
CREATE TABLE "ProcessedFor_PaymentTransaction_Merchant_EDG" (...);
CREATE TABLE "GovernedBy_PaymentTransaction_ComplianceRule_EDG" (...);
CREATE TABLE "HasAuditLog_PaymentTransaction_AuditLog_EDG" (...);
CREATE TABLE "HasRiskAssessment_PaymentTransaction_RiskAssessment_EDG" (...);

CREATE PROPERTY GRAPH "FIN_GRAPH"
  VERTEX TABLES (
    "PaymentTransaction_VTX" KEY ("transaction_id") LABEL "PaymentTransaction"
      PROPERTIES ("transaction_id","amount","currency","status",
                  "created_at","completed_at","reference_number"),
    "Account_VTX" KEY ("account_id") LABEL "Account"
      PROPERTIES ("account_id","account_type","balance","currency",
                  "status","opened_date","last_activity"),
    "Customer_VTX" KEY ("customer_id") LABEL "Customer"
      PROPERTIES ("customer_id","customer_type","name","email",
                  "phone","address","kyc_status"),
    "PaymentMethod_VTX" KEY ("method_id") LABEL "PaymentMethod" PROPERTIES (...),
    "Merchant_VTX" KEY ("merchant_id") LABEL "Merchant" PROPERTIES (...),
    "ComplianceRule_VTX" KEY ("rule_id") LABEL "ComplianceRule" PROPERTIES (...),
    "AuditLog_VTX" KEY ("log_id") LABEL "AuditLog" PROPERTIES (...),
    "RiskAssessment_VTX" KEY ("assessment_id") LABEL "RiskAssessment" PROPERTIES (...)
  )
  EDGE TABLES (
    "Initiates_Customer_PaymentTransaction_EDG" KEY ("id")
      SOURCE KEY ("src_Customer_id") REFERENCES "Customer_VTX"("customer_id")
      DESTINATION KEY ("dst_PaymentTransaction_id") REFERENCES "PaymentTransaction_VTX"("transaction_id")
      LABEL "Initiates" NO PROPERTIES,
    "FundsFrom_PaymentTransaction_Account_EDG" KEY ("id")
      SOURCE KEY ("src_PaymentTransaction_id") REFERENCES "PaymentTransaction_VTX"("transaction_id")
      DESTINATION KEY ("dst_Account_id") REFERENCES "Account_VTX"("account_id")
      LABEL "FundsFrom" NO PROPERTIES,
    "ReceivesTo_PaymentTransaction_Account_EDG" KEY ("id")
      SOURCE KEY ("src_PaymentTransaction_id") REFERENCES "PaymentTransaction_VTX"("transaction_id")
      DESTINATION KEY ("dst_Account_id") REFERENCES "Account_VTX"("account_id")
      LABEL "ReceivesTo" NO PROPERTIES,
    "AuthorizedBy_PaymentTransaction_PaymentMethod_EDG" KEY ("id")
      SOURCE KEY ("src_PaymentTransaction_id") REFERENCES "PaymentTransaction_VTX"("transaction_id")
      DESTINATION KEY ("dst_PaymentMethod_id") REFERENCES "PaymentMethod_VTX"("method_id")
      LABEL "AuthorizedBy" NO PROPERTIES,
    "ProcessedFor_PaymentTransaction_Merchant_EDG" KEY ("id")
      SOURCE KEY ("src_PaymentTransaction_id") REFERENCES "PaymentTransaction_VTX"("transaction_id")
      DESTINATION KEY ("dst_Merchant_id") REFERENCES "Merchant_VTX"("merchant_id")
      LABEL "ProcessedFor" NO PROPERTIES,
    "GovernedBy_PaymentTransaction_ComplianceRule_EDG" KEY ("id")
      SOURCE KEY ("src_PaymentTransaction_id") REFERENCES "PaymentTransaction_VTX"("transaction_id")
      DESTINATION KEY ("dst_ComplianceRule_id") REFERENCES "ComplianceRule_VTX"("rule_id")
      LABEL "GovernedBy" NO PROPERTIES,
    "HasAuditLog_PaymentTransaction_AuditLog_EDG" KEY ("id")
      SOURCE KEY ("src_PaymentTransaction_id") REFERENCES "PaymentTransaction_VTX"("transaction_id")
      DESTINATION KEY ("dst_AuditLog_id") REFERENCES "AuditLog_VTX"("log_id")
      LABEL "HasAuditLog" NO PROPERTIES,
    "HasRiskAssessment_PaymentTransaction_RiskAssessment_EDG" KEY ("id")
      SOURCE KEY ("src_PaymentTransaction_id") REFERENCES "PaymentTransaction_VTX"("transaction_id")
      DESTINATION KEY ("dst_RiskAssessment_id") REFERENCES "RiskAssessment_VTX"("assessment_id")
      LABEL "HasRiskAssessment" NO PROPERTIES
  );
\end{lstlisting}
\end{tcolorbox}

\clearpage
\section{Difficulty Taxonomy Examples}
\label{app:difficulty_taxonomy}

\Cref{tab:case_study_levels} provides concrete examples for each difficulty tier defined in Section~\ref{subsec:difficulty}. \textbf{Easy} queries involve only single-node or single-edge patterns with straightforward filtering and no aggregation. \textbf{Medium} queries extend to one-hop traversals with simple aggregation or basic filters. \textbf{Hard} queries introduce multi-hop or variable-length paths, multiple conditions, and non-nested aggregation. \textbf{Extra Hard} queries require multi-step \texttt{MATCH} clauses, nested logic, or complex path patterns, demanding both structural and reasoning depth.

\begin{table*}[h]
\centering
\caption{Case Study of Graph Query Complexity Taxonomy. The query logic evolves from easy single-node retrieval to extra-hard multi-step retrieval.}
\label{tab:case_study_levels}
\small
\setlength{\tabcolsep}{4pt}
\begin{tabularx}{\textwidth}{@{}l p{3.5cm} X@{}}
\toprule
\textbf{Difficulty} & \textbf{Characteristics} & \textbf{Case Example (Question \& Query)} \\
\midrule
\textbf{Easy}
& \textit{single node or edge pattern, no aggregation, no complex filtering}
& \textbf{Question:} Find accounts with account type \texttt{CREDIT}, return their opening date. \newline
    \texttt{\textbf{Query:} MATCH (n:ACCOUNT) WHERE n.account\_type = ``CREDIT'' RETURN n.opening\_date} \\
\midrule
\textbf{Medium}
& \textit{one-hop path, simple aggregation (e.g., \texttt{COUNT}, \texttt{SUM}) or basic filters, no nesting}
& \textbf{Question:} Retrieve alerts triggered by accounts with a risk score higher than 0.5, including alert severity level and the timestamp when the alert was created, limiting the results to 10. \newline
  \texttt{\textbf{Query:} MATCH (a:ACCOUNT)-[:TRIGGERED\_ALERT]-$>$(al:ALERT) WHERE a.risk\_score $>$ 0.5 RETURN al.severity\_level, al.creation\_timestamp LIMIT 10} \\
\midrule
\textbf{Hard}
& \textit{multi-hop paths ($\leq$2 hops or variable-length), multiple conditions, non-nested aggregation}
& \textbf{Question:} Find accounts reachable within 1 to 3 hops from a payment transaction with transaction\_id TXN0000000, limited to 5 results. \newline
  \texttt{\textbf{Query:} MATCH (n1:PaymentTransaction)-[]->\{1,3\}(n2:Account) WHERE n1.\newline transaction\_id = ``TXN0000000'' RETURN n2.account\_id, n2.account\_type LIMIT 5} \\
\midrule
\textbf{Extra Hard}
& \textit{complex paths ($\geq$3 hops), multi-step \texttt{MATCH}, nested aggregation, high structural and logical depth}
& \textbf{Question:} List all customers who have initiated at least one failed transaction and at least one completed transaction, showing their names and email addresses. \newline
  \texttt{\textbf{Query:} MATCH (c:Customer)-[:Initiates]-$>$(pt1:PaymentTransaction) WHERE pt1.\newline status = ``Failed'' RETURN c NEXT MATCH (c)-[:Initiates]-$>$\newline(pt2:PaymentTransaction) WHERE pt2.status = ``Completed'' RETURN DISTINCT c.name, c.email} \\
\bottomrule
\end{tabularx}
\end{table*}

\clearpage
\section{Error Analysis Case Study}
\label{app:error_analysis}

We present qualitative case studies for the four error categories identified in Section~\ref{subsec:multi_ql_perf}: \textit{Schema Linking Error}, \textit{Syntax Error}, \textit{Aggregation Error}, and \textit{Filter Error}. \Cref{tab:gql_case_study} shows GQL examples and \Cref{tab:sql_pgq_case_study} shows SQL/PGQ examples. Red text (\textcolor{red}{text}) highlights the erroneous parts in model predictions.

\begin{table*}[h]
    \centering
    \caption{Qualitative Case Study of Generation Failures in GQL.}
    \label{tab:gql_case_study}
    \small
    \renewcommand{\arraystretch}{1.3}
    \setlength{\tabcolsep}{4pt}
    \begin{tabular}{p{0.02\textwidth} p{0.22\textwidth} p{0.30\textwidth} p{0.40\textwidth}}
        \toprule
        \textbf{ID} & \textbf{Gold Standard} & \textbf{Model Prediction} & \textbf{Error Attribution} \\
        \midrule

        \textbf{\#1} &
        \makecell[l]{
            \textbf{Q:} How many distinct accounts \\
            have received funds from any \\
            payment transaction? \\
            \textbf{Gold:} \\
            \texttt{MATCH (t:PaymentTransaction)} \\
            \texttt{-\textbf{[:FundsFrom]}->(a:Account)} \\
            \texttt{RETURN COUNT(DISTINCT a)}
        } &
        \makecell[l]{
            \texttt{MATCH (:PaymentTransaction)} \\
            \texttt{-\textcolor{red}{[:ReceivesTo]}->(a:Account)} \\
            \texttt{RETURN count(DISTINCT a)}
        } &
        \textbf{Schema Linking Error}: The model uses \texttt{[:ReceivesTo]} (funds received into an account) instead of \texttt{[:FundsFrom]} (funds sourced from an account). These are two semantically opposite relationship types in the Financial Payment schema; using the wrong one entirely changes the query semantics. \\
        \midrule

        \textbf{\#2} &
        \makecell[l]{
            \textbf{Q:} Find all users who have \\
            posted tweets with media and \\
            are members of $\geq$2 lists. \\
            \textbf{Gold:} \\
            \texttt{... RETURN u NEXT MATCH} \\
            \texttt{(u)-[:MEMBER\_OF\_LIST]->(l)} \\
            \texttt{RETURN u, COUNT(l) AS c} \\
            \texttt{\textbf{NEXT} RETURN u.username, c}
        } &
        \makecell[l]{
            \texttt{... MATCH} \\
            \texttt{(u)-[:MEMBER\_OF\_LIST]->(l)} \\
            \texttt{\textcolor{red}{WITH} u, count(l) AS c} \\
            \texttt{\textcolor{red}{WHERE} c >= 2} \\
            \texttt{RETURN u.username, c}
        } &
        \textbf{Syntax Error}: The model places \texttt{WHERE} directly after \texttt{WITH}, adopting the Cypher-style \texttt{WITH\,...\,WHERE} pattern. ISO GQL requires \texttt{NEXT FILTER} or pre-filtering within \texttt{MATCH}. \\
        \midrule

        \textbf{\#3} &
        \makecell[l]{
            \textbf{Q:} Retrieve the average number \\
            of tweets per user for users \\
            who have posted $\geq$1 tweet \\
            and are members of $\geq$1 list. \\
            \textbf{Gold:} \\
            \texttt{... RETURN u, COUNT(t) AS c} \\
            \texttt{NEXT MATCH (u)-[:MEMBER...]} \\
            \texttt{RETURN AVG(c)}
        } &
        \makecell[l]{
            \texttt{... MATCH (u)-[:POSTS]->(t),} \\
            \texttt{(u)-[:MEMBER\_OF\_LIST]->(l)} \\
            \texttt{RETURN \textcolor{red}{AVG(COUNT(t))}}
        } &
        \textbf{Aggregation Error}: The model nests \texttt{COUNT(t)} inside \texttt{AVG(...)}, producing \texttt{AVG(COUNT(t))}. GQL explicitly forbids aggregation of aggregations; the gold query correctly separates the two steps via \texttt{NEXT}. \\
        \midrule

        \textbf{\#4} &
        \makecell[l]{
            \textbf{Q:} Retrieve all users who are \\
            either verified or have more \\
            than 100 followers. \\
            \textbf{Gold:} \\
            \texttt{MATCH (n:USER) WHERE} \\
            \texttt{n.verified = true} \\
            \texttt{\textbf{OR} n.followers > 100} \\
            \texttt{RETURN n.user\_id, n.username}
        } &
        \makecell[l]{
            \texttt{MATCH (u:USER) WHERE} \\
            \texttt{u.verified = true} \\
            \textcolor{red}{\texttt{AND}} \texttt{u.followers > 100} \\
            \texttt{RETURN u.user\_id, u.username}
        } &
        \textbf{Filter Error}: The gold query uses \texttt{OR} (union---verified users or users with $>$100 followers), but the predicted query replaces it with \texttt{AND} (intersection---both conditions must hold). This logical operator reversal fundamentally changes the result set. \\
        \bottomrule
    \end{tabular}
\end{table*}

\begin{table*}[t]
    \centering
    \caption{Qualitative Case Study of Generation Failures in ISO SQL/PGQ.}
    \label{tab:sql_pgq_case_study}
    \small
    \renewcommand{\arraystretch}{1.3}
    \setlength{\tabcolsep}{4pt}
    \begin{tabular}{p{0.02\textwidth} p{0.22\textwidth} p{0.30\textwidth} p{0.40\textwidth}}
        \toprule
        \textbf{ID} & \textbf{Gold Standard} & \textbf{Model Prediction} & \textbf{Error Attribution} \\
        \midrule

        \textbf{\#1} &
        \makecell[l]{
            \textbf{Q:} Retrieve distinct bio categories \\
            from all USER nodes. \\
            \textbf{Gold:} \\
            \texttt{SELECT DISTINCT * FROM} \\
            \texttt{GRAPH\_TABLE ("TW\_GRAPH"} \\
            \texttt{MATCH (n IS "USER")} \\
            \texttt{COLUMNS (n.\textbf{"bio"} AS bio)) gt}
        } &
        \makecell[l]{
            \texttt{SELECT DISTINCT bio FROM} \\
            \texttt{GRAPH\_TABLE (TW\_GRAPH} \\
            \texttt{MATCH (u IS "USER")} \\
            \texttt{COLUMNS (\textcolor{red}{u.bio} AS bio))}
        } &
        \textbf{Schema Linking Error}: The predicted query refers to \texttt{u.bio} without double quotes. Oracle consequently resolves it as the uppercase identifier \texttt{BIO}, rather than the case-sensitive property \texttt{"bio"}, and raises ORA-40983. \\
        \midrule

        \textbf{\#2} &
        \makecell[l]{
            \textbf{Q:} List all distinct usernames \\
            from the USER nodes. \\
            \textbf{Gold:} \\
            \texttt{SELECT DISTINCT * FROM} \\
            \texttt{GRAPH\_TABLE ("TW\_GRAPH"} \\
            \texttt{MATCH (u \textbf{IS} "USER")} \\
            \texttt{\textbf{COLUMNS} (u."username" ...) gt}
        } &
        \makecell[l]{
            \texttt{SELECT DISTINCT u.username} \\
            \texttt{FROM GRAPH\_TABLE(} \\
            \textcolor{red}{\texttt{'TW\_GRAPH'}} \\
            \texttt{MATCH (u\textcolor{red}{:USER})) AS u}
        } &
        \textbf{Syntax Error}: The model treats the graph name as a string literal, uses the Cypher label form \texttt{u:USER} instead of \texttt{u IS "USER"}, omits the required \texttt{COLUMNS} clause, and uses an invalid alias form. Oracle raises ORA-03050. \\
        \midrule

        \textbf{\#3} &
        \makecell[l]{
            \textbf{Q:} Find the total value of \\
            contracts from vendors with \\
            rating $<$3.5. \\
            \textbf{Gold:} \\
            \texttt{SELECT name, SUM(value)} \\
            \texttt{FROM GRAPH\_TABLE (...} \\
            \texttt{COLUMNS (v."name", c."value")} \\
            \texttt{) gt GROUP BY name}
        } &
        \makecell[l]{
            \texttt{SELECT v\_name, totalVal} \\
            \texttt{FROM GRAPH\_TABLE (...} \\
            \texttt{COLUMNS (v."name",} \\
            \texttt{\textcolor{red}{SUM(c."value")} AS totalVal)} \\
            \texttt{) gt GROUP BY v\_name}
        } &
        \textbf{Aggregation Error}: The model evaluates \texttt{SUM(c."value")} inside the \texttt{COLUMNS} clause of \texttt{GRAPH\_TABLE}. The property should be projected from \texttt{GRAPH\_TABLE} and then aggregated by the outer SQL query. Oracle rejects this with ORA-49011. \\
        \midrule

        \textbf{\#4} &
        \makecell[l]{
            \textbf{Q:} Get the product SKU and \\
            quantity for work order \\
            WO00000, limited to 10. \\
            \textbf{Gold:} \\
            \texttt{SELECT * FROM GRAPH\_TABLE} \\
            \texttt{(... MATCH (n IS "WorkOrder")} \\
            \texttt{\textbf{WHERE n."WorkOrder\_id"}} \\
            \textbf{= 'WO00000'} \texttt{COLUMNS (...) gt} \\
            \texttt{FETCH FIRST 10 ROWS ONLY}
        } &
        \makecell[l]{
            \texttt{SELECT * FROM GRAPH\_TABLE} \\
            \texttt{(... MATCH (wo IS "WorkOrder")} \\
            \texttt{COLUMNS (...)) gt} \\
            \textcolor{red}{\texttt{WHERE wo."WorkOrder\_id"}} \\
            \textcolor{red}{\texttt{= 'WO00000'}} \\
            \texttt{FETCH FIRST 10 ROWS ONLY}
        } &
        \textbf{Filter Error}: The filter is placed outside \texttt{GRAPH\_TABLE}, where the graph variable \texttt{wo} is no longer in scope and \texttt{WorkOrder\_id} was not projected as a relational column. The predicate belongs inside \texttt{GRAPH\_TABLE}. Oracle raises ORA-00904. \\
        \bottomrule
    \end{tabular}
\end{table*}

\clearpage
\clearpage
\section{Cross-Query-Language Transfer between SQL and Cypher}

\begin{figure}[!htb]
    \centering
    \includegraphics[width=\textwidth]{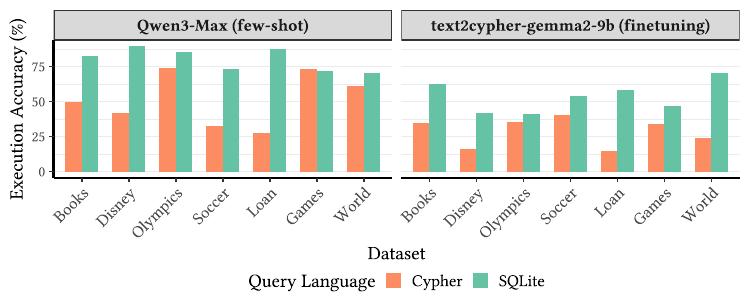}
    \caption{Cross-Query-Language Transfer Performance Comparison between SQL and Cypher on BIRD-derived domains.}
    \label{fig:bird_cross_lingual}
\end{figure}

We evaluate cross-query-language generalization using Qwen3-Max under few-shot prompting and text2cypher-gemma2-9b under fine-tuning on BIRD-derived test domains.
\Cref{fig:bird_cross_lingual} reports execution accuracy when transferring between SQLite and Cypher across seven domains.
Across all domains and both models, performance on SQLite consistently exceeds that on Cypher, revealing a persistent asymmetry in cross-lingual transfer. This gap remains evident both under few-shot prompting and task-specific fine-tuning, suggesting that limited exposure to target-language examples is insufficient to fully bridge query language differences.
The transfer gap is strongly domain-dependent: Qwen3-Max drops sharply on \textit{Loan} and \textit{Disney} but remains comparatively stable on \textit{Olympics} and \textit{Games}.
text2cypher-gemma2-9b exhibits the same trend with lower overall accuracy, suggesting that fine-tuned models are more sensitive to query language shifts. These results indicate that cross-lingual generalization is shaped by both query language syntax and domain-specific query structure inherited from relational sources.

\section{Ethics and Fairness}

We discuss ethical considerations along four dimensions.

\noindent\textbf{Data Privacy.}
All seed queries in the synthesis pipeline are derived from real business queries through a desensitization and abstraction process: proprietary identifiers, customer information, and sensitive business logic are removed or replaced with synthetic equivalents before any data enters the benchmark. The generated graph databases contain only fictitious entities produced by controlled data generators; no real user data or production records are included.

\noindent\textbf{Consent and Licensing.}
Existing datasets incorporated into the benchmark (e.g., Text2Cypher, FinBench, BIRD) are used under their original licenses and terms of use. For synthesized domains, schema specifications and seed queries are authored by domain experts who have consented to their inclusion in the benchmark. The benchmark itself is released under a permissive license that permits academic use while prohibiting re-identification attempts.

\noindent\textbf{Bias.}
The benchmark covers 13 domains selected to span both well-resourced areas (e.g., Social Network, Finance) and underrepresented ones (e.g., Manufacturing, IT Asset Management). Nevertheless, domain coverage is not uniform, and the distribution of query complexity may reflect the characteristics of the source datasets. We encourage users to interpret evaluation results with this skew in mind and to avoid over-generalizing performance claims to unseen domains.

\noindent\textbf{Potential Misuse.}
The benchmark is intended solely for evaluating and improving text-to-graph-query systems. We explicitly discourage using the synthetic data or query patterns to reverse-engineer proprietary business logic, to train systems for unauthorized data access, or to benchmark against production databases without proper authorization.

\end{document}